\documentclass{article}
\usepackage{iclr2027_conference,times}
\iclrfinalcopy
\usepackage[T1]{fontenc}
\usepackage{microtype,amsmath,amssymb,booktabs,tabularx,longtable,array}
\usepackage{graphicx,xcolor,enumitem,needspace,zref-savepos,zref-abspage}
\usepackage{hyperref}

\hypersetup{colorlinks=true,linkcolor=reportgreen,urlcolor=reportgreen,citecolor=reportgreen,
 pdftitle={AgentEvolver: System-Wide Self-Evolution Through Task Execution},
 pdfauthor={AgentEvolver Project},pdfsubject={System-wide self-evolution through task execution}}
\definecolor{reportgreen}{HTML}{32755C}
\definecolor{reportmint}{HTML}{ECFCF6}
\definecolor{reportgray}{HTML}{5B6E67}
\definecolor{reportamber}{HTML}{7D6640}
\definecolor{reportink}{HTML}{0E1C19}
\definecolor{reportblue}{HTML}{4C6E80}
\definecolor{reportpurple}{HTML}{6A6686}
\definecolor{reportgold}{HTML}{7D6640}
\definecolor{reportline}{HTML}{D1D7D5}
\definecolor{reportpaper}{HTML}{F9F9F9}
\newcolumntype{Y}{>{\raggedright\arraybackslash}X}
\newcolumntype{L}[1]{>{\raggedright\arraybackslash}p{#1}}
\makeatletter
\zref@addprops{savepos}{abspage}
\makeatother
\newcommand{\para}[2]{\par\zsavepos{para:#1:start}#2\zsavepos{para:#1:end}\par}
\newcommand{\code}[1]{\texttt{\small #1}}

\title{AgentEvolver: System-Wide Self-Evolution\\Through Task Execution}
\author{Wentao Zhang\textsuperscript{1}, Fuchao Yang\textsuperscript{1}, Yilei Zhao\textsuperscript{1}, Xinrun Wang\textsuperscript{2}, Bo An\textsuperscript{1}\\
\normalfont\textsuperscript{1}Nanyang Technological University, Singapore\\
\normalfont\textsuperscript{2}Singapore Management University, Singapore\\
\normalfont\texttt{zhangwent963@gmail.com}\\
\url{https://dvampire.github.io/AgentEvolver/}\\
\url{https://github.com/DVampire/AgentEvolver}}
\begin{document}
\maketitle
\fancyhead{}

\begin{abstract}
\para{abstract}{An agent can complete a task without improving how it works. Turning task experience into reusable capability requires connecting the changed component to its evaluation and subsequent use. We present AgentEvolver, a system for developing capabilities during task execution while keeping the foundation model fixed. Eight entity families expose reusable operations, methods, agents, control flow, interfaces, and supporting state to revision through a common versioned lifecycle. A shared Runtime coordinates ongoing work, while persistent planning and recoverable context preserve task direction and supporting evidence. We evaluate task outcomes on SWE-bench Pro Public and examine capability changes in six application cases. The team reports an 82.08\% resolution rate with evolution, exceeding its reported baseline without evolution. The cases show retained capabilities entering later website, game, and research work, while also documenting incomplete objectives and an unsuccessful strategy. These findings distinguish improvement in a reusable component from success on the final task. AgentEvolver provides a concrete basis for studying capability accumulation through execution; independent-task transfer and total development cost remain open questions.}
\end{abstract}

\section{Introduction}
\label{sec:introduction}
\begin{figure}[t]
\centering
\includegraphics[width=\linewidth]{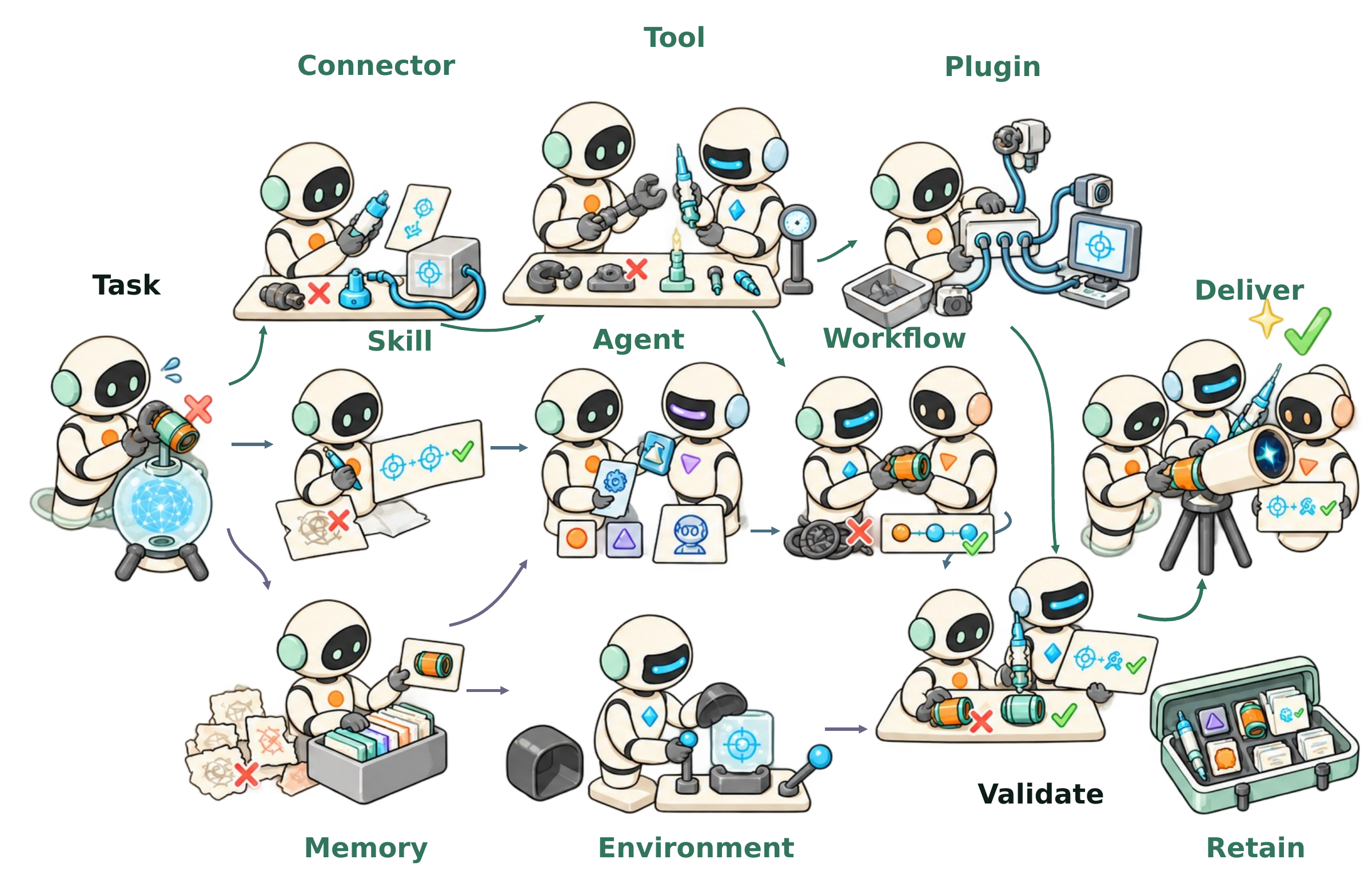}
\caption{\textbf{Task-driven capability evolution.} An observatory repair motivates complementary changes across eight entity families. Dependencies between branches lead to shared evaluation, collaborative delivery, and retained capabilities. Plan preserves the goal while Runtime coordinates the work. The scene is a conceptual analogy; revised objects and edges describe this illustrative task rather than a robotics experiment.}
\label{fig:story}
\end{figure}

\para{intro:story}{Imagine a mind in a glass vessel that knows how a telescope works but cannot adjust its lens. Given a robot body and an observatory to restore, it can act, yet its first image is blurred. A better probe alone will not solve the problem: the instrument interface must supply the right measurements, the calibration method must use them correctly, and specialist robots must coordinate their work. The assignment gradually changes both the observatory and the system repairing it. Figure~\ref{fig:story} illustrates how the branches connect an initial failure to a tested revision and collaborative delivery.}

\para{intro:storygrowth}{The story leaves two outputs: a repaired observatory and capabilities that can be used again. Software agents face a similar opportunity. Fixing a repository can reveal a missing diagnostic operation; revising a website can expose an inadequate review method; testing a hypothesis can require a more reproducible research environment. A useful response may therefore change the current deliverable, the agent's means of working, or both. This motivates our research question: \emph{how can task execution develop reusable capabilities with inspectable evidence of evaluation and subsequent use?}}

\para{intro:prior}{Existing agents already turn interaction into retained feedback and executable skills \citep{shinn2023reflexionlanguageagentsverbal,wang2023voyageropenendedembodiedagent}. Agent-program search and modification during execution further make the agent's own software an object of adaptation \citep{hu2025automateddesignagenticsystems,zhang2026darwingodelmachineopenended,xia2025livesweagentsoftwareengineeringagents}. These developments motivate a systems question: how should heterogeneous revisions coexist with an unfinished task, its resource budget, and the evidence supporting earlier decisions? AgentEvolver focuses on the interfaces and execution mechanisms needed to maintain those connections.}

\para{intro:challenge}{The challenge is maintaining consistency while work and capabilities change together. A revised interface may invalidate a specialist's assumptions; an evaluation may refer to a version that has since been replaced; a context summary may omit why a candidate was rejected. These problems become coupled when agents delegate work and share finite resources. Addressing them requires more than generating an additional tool. The system must connect the identity of a change, the execution that tests it, and the task state in which it is subsequently consumed.}

\para{intro:approach}{AgentEvolver links a candidate's identity, evaluation, retention, and later use through a versioned lifecycle. The acting agent proposes and judges revisions; Runtime coordinates the work under shared ownership and resource accounting. A persistent Plan and recoverable Context preserve the objective and its evidence as the interaction grows. Tasks requiring verified improvement additionally require a successful later consumer. Together, these mechanisms connect capability development to the continuing task.}

\para{intro:global}{We call the supported scope \emph{global capability evolution}: Tool, Skill, Agent, Workflow, Connector, Environment, Memory, and Plugin can change while the foundation model remains fixed. Global refers to the working system's adaptation surfaces, rather than an optimal search guarantee. The contribution is the coordination of heterogeneous changes under shared execution and evidence requirements. A task can use only the families it needs, and support for a family does not itself establish an empirical improvement in that family.}

Our contributions are threefold:
\begin{itemize}
\item \textbf{A system model for capability evolution during task execution.} Eight heterogeneous component families share version identities and an evidence lifecycle linking construction, evaluation, retention, and use.
\item \textbf{Continuity across task and capability changes.} A shared Runtime, persistent planning, and recoverable context management coordinate capability development with the original objective, its dependencies, and its resource budget.
\item \textbf{A two-level empirical study.} A SWE-bench Pro Public comparison and six application cases examine resolved tasks, component changes, and later consumers, including incomplete objectives and a negative strategy result.
\end{itemize}

\para{intro:results}{The empirical studies examine both outputs of this process. The team-reported SWE-bench Pro Public result rises from 79.07\% to 82.08\% with evolution enabled. Six historical application cases then examine how particular changes were evaluated and used, including a chain linking sources, design, and review, a faster engine-diagnostic operation, and reproducible research environments. Partial and unsuccessful outcomes reveal where a retained capability is insufficient for final task acceptance. Together, the coding results and case reconstructions examine task outcomes and the changes that support them.}

\para{intro:roadmap}{Sections~\ref{sec:related} to~\ref{sec:evaluation} develop the positioning, system, and empirical studies. Section~\ref{sec:discussion} examines validity and remaining questions. Appendices~\ref{app:core} to~\ref{app:protocol} provide mechanism details, case reconstructions, and evaluation protocols that support the analysis.}

\section{Related Work}
\label{sec:related}
\label{sec:positioning}
\para{related:axis}{Prior work makes experience reusable through several mechanisms: feedback stored for another attempt, executable skills, revised agent programs, and learned model parameters. We organize the comparison around the object of adaptation and the stage at which it changes. AgentEvolver focuses on heterogeneous software components revised during an unfinished task, with a common connection between execution, version identity, and retention evidence.}

\subsection{Feedback and reusable capabilities}
\para{related:feedback}{ReAct \citep{yao2023reactsynergizingreasoningacting} interleaves reasoning and action with environmental observations. Reflexion \citep{shinn2023reflexionlanguageagentsverbal} uses retained verbal feedback to influence later attempts, while Voyager \citep{wang2023voyageropenendedembodiedagent} develops an executable skill library through exploration and refinement in Minecraft. These approaches demonstrate that useful experience can reside outside model parameters. AgentEvolver makes the retained object explicit: a task record, an operating method, and an executable operation have different interfaces and evaluation requirements. The component model preserves these distinctions within a common process of development, evaluation, and use.}

\para{related:tools}{Toolformer \citep{schick2023toolformerlanguagemodelsteach} trains a language model to select and use APIs through self-supervised examples. This changes the model's tool-use behavior. AgentEvolver instead holds the foundation model fixed and permits changes to the surrounding capability system. The distinction concerns the object being adapted: learning when to call an existing API and revising the operation or method exposed to the agent require different evaluations.}

\subsection{Agent design and runtime self-modification}
\para{related:search}{DSPy \citep{khattab2023dspycompilingdeclarativelanguage} optimizes modular language-model pipelines, while Automated Design of Agentic Systems \citep{hu2025automateddesignagenticsystems} searches over agent programs using an archive and evaluation feedback. The Darwin G\"odel Machine \citep{zhang2026darwingodelmachineopenended} develops an expanding archive of modified coding agents. These approaches place different parts of agent software within an optimization process. AgentEvolver emphasizes the infrastructure in which heterogeneous revisions are constructed, evaluated, and consumed during a continuing task. The acting agent supplies the proposal policy, while the framework maintains component identities and evidence throughout development.}

\para{related:live}{Live-SWE-agent \citep{xia2025livesweagentsoftwareengineeringagents} modifies its scaffold while solving software-engineering tasks, making it particularly close in adaptation timing. AgentEvolver studies the coordination requirements of this setting across a broader collection of component interfaces. Process ownership, recoverable task state, and version-specific observations allow several changes to participate in the same continuing task. The contribution concerns how those changes are managed and examined. These experiments do not establish superiority over Live-SWE-agent in a controlled comparison.}

\para{related:hyper}{Hyperagents \citep{zhang2026hyperagents} combine a task agent and the meta-level modification procedure within an editable program. This extends adaptation to the mechanism that generates future revisions. AgentEvolver's present contribution is complementary: it defines how component changes are executed, identified, assessed, and reused. Its eight-family scope does not establish recursive improvement of the revision procedure; that claim requires a separate longitudinal comparison.}

\para{related:learning}{Self-questioning, self-navigating, and self-attributing are the three mechanisms studied in AgentEvolver: Towards Efficient Self-Evolving Agent System \citep{zhai2025agentevolverefficientselfevolvingagent}. Its emphasis is task generation, exploration, and differentiated reward assignment. The present report examines versioned software capabilities during execution with a fixed foundation model. These approaches change different parts of an agent, so their evaluation must identify the object of adaptation.}

\subsection{Coordination, context, and evaluation}
\para{related:coordination}{AutoGen \citep{wu2023autogenenablingnextgenllm} organizes configurable multi-agent conversations, while MetaGPT \citep{hong2024metagptmetaprogrammingmultiagent} structures collaboration through roles and operating procedures. These systems establish coordination as an explicit design concern. AgentEvolver connects coordination to capability development: Runtime associates processes and operations with owners, resources, and observable lifecycle states. These mechanisms preserve the relation between delegated work and the candidate version it produces. Ablations are needed to quantify the contribution of each mechanism.}

\para{related:context}{Generative Agents \citep{park2023generativeagentsinteractivesimulacra} combine memory, reflection, and planning to support continuing behavior. MemGPT \citep{packer2024memgptllmsoperatingsystems} manages information across model context and external storage. Evidence that retrieval performance depends on where information appears in a long context \citep{liu2023lostmiddlelanguagemodels} further motivates evaluating what the agent can recover, rather than considering context length alone. AgentEvolver separates the current Plan, working context, and archived observations so that the agent can recover both its task direction and the observations supporting a capability decision.}

\subsection{Task evaluation and adaptive selection}
\para{related:swe}{SWE-bench \citep{jimenez2024swebenchlanguagemodelsresolve} evaluates repository issue resolution, SWE-agent \citep{yang2024sweagentagentcomputerinterfacesenable} examines the agent-computer interface, and SWE-bench Pro \citep{deng2025swebenchproaiagents} extends evaluation to challenging software-engineering tasks. Task identity, available feedback, resource limits, and grading conditions govern what comparisons support. We distinguish our reported internal intervention from external reference values under different conditions. Repeated selection on observed feedback can also bias an evaluation \citep{JMLR:v11:cawley10a,dwork2015generalizationadaptivedataanalysis}; our case analysis therefore separates development comparisons, later use, and independent held-out assessment.}

\section{AgentEvolver}
\label{sec:system}
\subsection{Problem setting and design requirements}
\label{sec:setup}
\para{setup:model}{Consider a task $q$ with a deliverable criterion, an initial capability collection, and a finite resource budget. The foundation model remains fixed. An execution changes both the requested artifact and the reusable means of producing it. To describe these changes, we distinguish six parts of the execution state:
\begin{equation}
S_t=(D_t,\mathcal{L}_t,A_t,R_t,H_t,b_t).
\label{eq:state}
\end{equation}
Here $D_t$ is the deliverable state; $\mathcal{L}_t$ is the archive of admitted component versions; $A_t$ identifies active versions; $R_t$ records processes, invocations, and ownership; $H_t$ contains planning and interaction records; and $b_t$ is the remaining budget. This notation separates the historical record of a revision from the version currently available for task execution.}

\para{setup:goals}{Ordinary task actions operate on $D_t$ using capabilities selected through $A_t$. Evolution actions can add an archived version to $\mathcal{L}_t$, change the active selection, or append evaluation and decision records to $H_t$. Both kinds of work share Runtime and budget accounting. A candidate may become provisionally active before functional evaluation, so archive membership, current availability, and justified retention are distinct properties. The final task is assessed against its deliverable criterion; a capability claim is assessed against a named version, comparison conditions, and observed consumers.}

\begin{figure}[t]
\centering
\includegraphics[width=\linewidth]{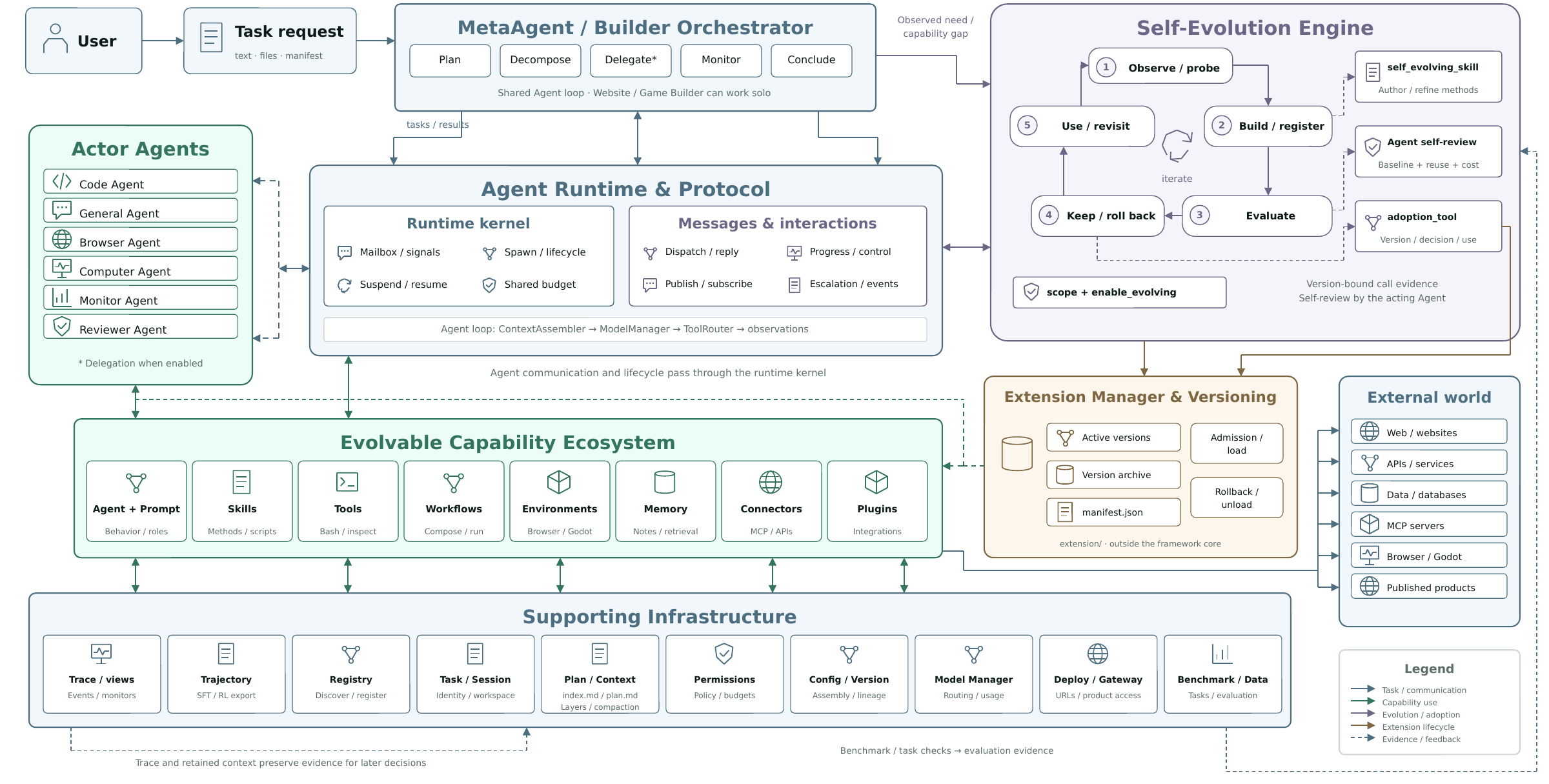}
\caption{\textbf{AgentEvolver system architecture.} The Runtime connects task orchestration, eight capability families, and the self-evolution loop. Version management links evaluated changes to active capabilities; shared infrastructure supports planning, context, and evidence. Arrows distinguish execution, capabilities, evolution, versioning, and feedback.}
\label{fig:architecture}
\end{figure}

\para{setup:scope}{Three design requirements follow. \textbf{Identity} must survive later edits so that an evaluation continues to refer to the component that was tested. \textbf{Continuity} must preserve task direction and execution ownership while the agent develops a candidate. \textbf{Evidence} must connect a retention decision to actual observations and subsequent use. The eight-family model, Runtime, Plan/Context, and evolution lifecycle implement these requirements together (Figure~\ref{fig:architecture}). Appendix~\ref{app:core} explains their detailed semantics and the responsibilities of each mechanism.}

\subsection{Eight evolution surfaces and the unit of change}
\para{system:entities}{Table~\ref{tab:entities} summarizes the eight families. Tool, Skill, Agent, Workflow, Connector, and Plugin participate in the callable capability roster. Environment supplies mounted state and actions, and Memory is supplied internally through its owning subsystem. Their shared identity is a tuple $(f,n,v)$: family, registered name, and version. Common discovery and versioning let an agent find and revise capabilities without requiring every family to behave like a stateless function.}

\begin{table}[t]
\caption{Eight component families: reusable change surfaces and operational roles.}
\label{tab:entities}
\small
\begin{tabularx}{\linewidth}{@{}lYY@{}}
\toprule
Family & Reusable change surface & Operational role \\
\midrule
Tool & Executable operation and input schema & Bounded call through the tool pipeline \\
Skill & Method, checks, scripts, and references & Instructions loaded for an agent to apply \\
Agent & Behavior, prompt, and configuration & Runtime process with a mailbox and lifecycle \\
Workflow & Structured executable control flow & Managed execution with frames and checkpoints \\
Connector & Provider-facing actions and contracts & Data or service access through named operations \\
Environment & Observations, state, and action interface & Mounted task environment \\
Memory & Retention and retrieval implementation & Internally supplied context backend \\
Plugin & Service implementation and related tools & Tool container with shared client and configuration \\
\bottomrule
\end{tabularx}
\end{table}

\para{system:consumer}{The appropriate change surface depends on the consumer. A bounded operation suggests a Tool; a procedure requiring judgment across observations suggests a Skill. A specialist with separate instructions and context suggests an Agent, while reusable control flow suggests a Workflow. Provider access and shared services motivate Connector and Plugin changes. Environment and Memory change the system's observation or retention mechanisms. These interfaces allow a task to combine revisions without requiring every component to behave like a stateless function.}

\para{system:rolechange}{The component definition, its instances, and its stored content must remain distinguishable. Dispatching another worker from an unchanged definition creates an Agent process; revising its instructions or mounted capabilities changes the reusable Agent. An Agent prompt therefore belongs to that family rather than forming a ninth entity. Likewise, writing a fault record is a Memory-content update, while revising its indexing or retrieval implementation changes the Memory component. Updating today's Plan differs from changing a reusable Workflow. These distinctions prevent ordinary task activity from being counted automatically as capability evolution.}

\para{system:versions}{The extension layer archives the bytes admitted for each version and rejects different content submitted under an existing immutable version. Family-specific managers remain responsible for loading the component and its domain state. A Skill retains its manifest, scripts, and references; a service Plugin can retain shared clients and configuration; a Workflow retains its control structure. Version identities make these heterogeneous artifacts addressable by evaluations and later consumers, while preserving the representation appropriate to each family.}

\subsection{Runtime: coordinating work while capabilities change}
\para{system:process}{Capability development must coexist with the work that motivated it. Runtime separates persistent Agent processes from bounded component invocations. A process owns its conversation, mailbox, and child work; a component invocation records a particular operation and its result. Workers can complete an assignment, resident services can accept further requests, and subscribers can react to addressed topics. Explicit lifecycle states let a coordinator distinguish completed work from waiting or suspended work before deciding whether a revision is ready to use.}

\para{system:invocation}{Operations pass through a common invocation boundary that records their owner, parent call, component identity, and execution status. Declared shared and exclusive resource claims control conflicting access. This matters when one branch revises an artifact that another branch is inspecting: concurrency should follow available independence information. The common boundary also lets evaluation records connect observations to the executed component version.}

\para{system:cancellation}{Ownership continues through interruption and cleanup. A parent must stop and join its children before completing its own exit, and an in-flight synchronous operation retains its resource claims until it finishes. Suspension occurs at defined boundaries between steps. These rules keep unfinished activity visible when a task pauses or changes direction. They coordinate declared operations; uncertain external effects still require reconciliation after an interruption.}

\para{system:budgets}{The root run and its descendants share resource accounting, including later turns of resident processes. Candidate construction, failed attempts, evaluation, and actual consumption therefore belong to the cost of the original task. Reported, estimated, and unknown usage remain distinguishable. This accounting supports a meaningful comparison between evolving and fixed configurations: a faster retained Tool can still be expensive to obtain, and reporting only its final invocation would omit that development cost. Appendix~\ref{app:contracts} details these execution contracts.}

\subsection{Plan and Context: preserving direction and evidence}
\para{system:plan}{Planning and memory support continuing agent behavior \citep{park2023generativeagentsinteractivesimulacra}. In our setting, a capability revision is a subproblem within the original assignment. Plan preserves that relationship through a compact current index linked to detailed records of objectives, acceptance criteria, dependencies, and decisions. The latest index is included before model requests; supporting records are retrieved when needed. This division keeps immediate direction accessible without repeatedly placing every historical detail in the working context. A completed plan entry expresses the agent's judgment, while deliverable acceptance remains a separate evaluation.}

\para{system:context}{Hierarchical context management separates the material immediately available to a model from information retained in external storage \citep{packer2024memgptllmsoperatingsystems}. AgentEvolver combines stable instructions, a checkpoint summary, and recent complete interactions. Updated Plan and environment observations enter the conversation before request assembly. When older interactions are compacted, the system retains an archive and checks whether the proposed summary reduces the complete request; otherwise it restores the original history. Recent cycles of calls and results remain intact. The design reduces working-context pressure and preserves access to omitted details. The agent must still retrieve and assess the evidence relevant to its next decision.}

\para{system:planmodes}{Programmatic tool use provides a complementary control over context growth. An agent can compose several operations and return selected results through the same invocation boundary, retaining ownership and relationships between parent and child calls. This allows a diagnostic procedure to inspect intermediate artifacts without presenting every intermediate value to the model. Planning modes and permission checks apply at the same boundary. Appendix~\ref{app:planning} describes how these controls govern task execution.}

\para{system:trace}{Trace records the effective model request, invoked actions, and returned observations. Together with the version archive, these records allow a retention decision to be reconstructed after the conversation has moved on. Plan supplies direction, Context supplies the material for the next decision, and Trace preserves what was actually executed. The three mechanisms serve different roles in maintaining continuity across task and capability changes.}

\subsection{From an observed gap to a retained capability}
\label{sec:evolution}
\para{evolution:gap}{The acting agent chooses when and how to propose a change under the task's instructions and remaining budget. A proposal identifies a consumer need, the current limitation, and observations that expose it. The acceptance criterion is domain-dependent: an interface may need aligned dates, while a diagnostic operation may need identical outputs at lower latency. Verified-improvement tasks require baseline observations before registration. The system supplies the lifecycle and evidence checks; the agent supplies the proposal policy and substantive comparison criterion.}

\begin{figure}[t]
\centering
\includegraphics[width=\linewidth]{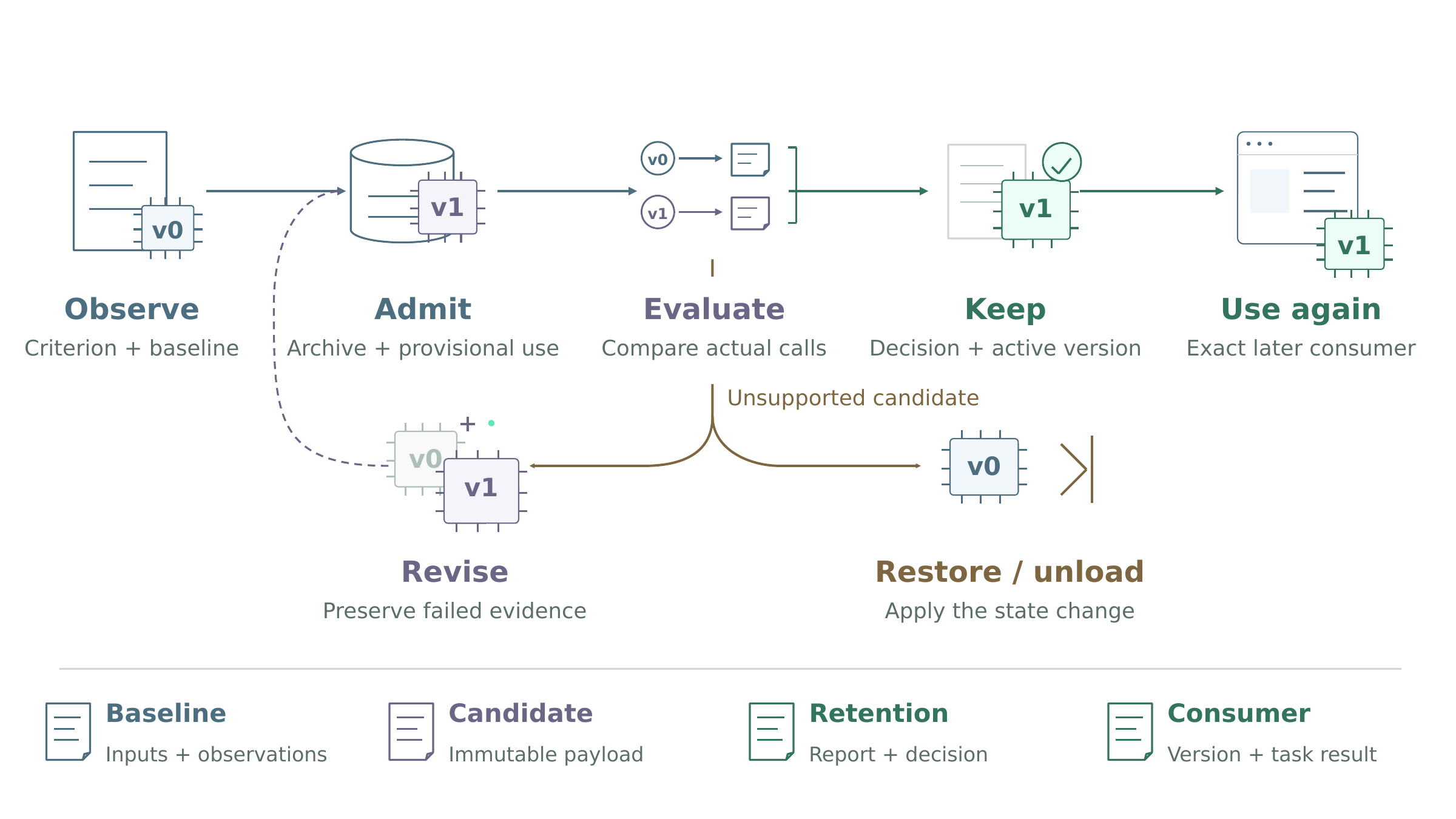}
\caption{\textbf{Component evolution and its evidence.} Structural admission, functional evaluation, retention, and later use are distinct steps. Unsupported candidates preserve failure evidence and return for revision or an explicit restore/unload operation. The lower rail associates observations with component identity; it is schematic, not a measured version trajectory. Tool-specific shadow/canary mechanisms supplement this shared lifecycle.}
\label{fig:lifecycle}
\end{figure}

\para{evolution:admission}{The candidate is authored in its family's representation and submitted for structural admission. A separate isolated interpreter checks loading, construction, and schema requirements; admitted bytes are archived under their version. Registration may provisionally activate the candidate before functional evaluation. This ordering separates two questions: whether the candidate can enter the system and whether its observed behavior justifies retention. Figure~\ref{fig:lifecycle} shows the lifecycle, and Table~\ref{tab:algorithm} summarizes the corresponding operations.}

\para{evolution:evaluation}{Functional evaluation names the exact family, component, and version, describes the baseline, and records expected and observed results for nonempty test cases. Each case references executed calls. The adoption Tool checks those references against retained observations, and a passing report requires every case to pass. The submitting agent supplies the substantive judgment; the framework checks provenance and internal consistency. Independent tests, measurements, and external graders can strengthen the evidence, while a poor fixture or an unfair baseline remains a limitation even when the call references are valid.}

\para{evolution:decision}{A keep decision requires the evaluated version to match the active archived version and to carry passing evidence with the required run fields. Decisions are appended to history. Rollback and unload have separate state-changing operations, so a review must inspect the resulting active state as well as the recorded intention. Failed candidates and replaced versions remain part of the development history. Retention thus records a revisable engineering decision with observed scope, rather than declaring a component universally reliable.}

\para{evolution:use}{Actual use answers a further question: did the retained capability participate in the work? Verified-improvement tasks require a subsequent successful consumer tied to the retained version, together with the task's comparison or reuse evidence. For a Skill, loading the instructions is only the first step; the observations must show the method being applied. Final deliverable acceptance is evaluated separately. Admission, functional evaluation, adoption, and later use are therefore distinct events. Evidence for one does not establish that the others occurred.}

\begin{table}[t]
\caption{Task-driven evolution protocol. The acting agent proposes candidates and comparison criteria; the framework manages version identity, call records, and retention checks. The task instruction determines any additional consumer requirement.}
\label{tab:algorithm}
\small
\begin{tabularx}{\linewidth}{@{}rY@{}}
\toprule
Step & Action and observable condition \\
\midrule
1 & Observe a task limitation; retain the baseline calls and define a consumer-facing acceptance criterion. \\
2 & Author a candidate and register its exact family, name, and version. If admission fails, retain the failure and revise or abandon the candidate. \\
3 & Evaluate the admitted candidate against the baseline and relevant regression cases; collect actual call identifiers and observed results. \\
4 & If evidence is missing, the version mismatches, or a required case fails, do not record a passing keep. Revise, restore a prior version, or unload as appropriate. \\
5 & Record a supported decision and execute any required state-changing operation separately. Preserve the archived payload and decision history. \\
6 & Use the retained version in subsequent task work. Record the consumer and resulting artifact; evaluate the final deliverable independently. \\
\bottomrule
\end{tabularx}
\end{table}

\para{evolution:cross}{The interfaces compose along task dependencies. In LUMEN, a source Connector emits an evidence packet, a specialist Agent authors an inquiry from that packet, and a review Skill checks the route's claims and measurements. Revising the source contract can therefore require corresponding changes to authoring and review. Version identities and call records make the combined chain inspectable, while the domain criterion determines whether its output is acceptable. This chain illustrates the method in a case whose component families were prescribed by the task.}

\para{evolution:comparison}{The comparison itself is domain-dependent. An astronomy interface must preserve relevant dates and reference frames; a voyage review must control the saved starting state; a timing comparison must retain the same input workload. A retained version should carry these applicability conditions forward to its consumers. Reuse can then reveal another limitation and motivate a new revision. This closes the task-level development loop while reserving independent-task transfer for an evaluation that freezes the capability library before exposing new consumer feedback.}

\section{Empirical Studies}
\label{sec:evaluation}

\noindent\textbf{Study design.}\par
\para{eval:questions}{We ask two complementary questions. \textbf{RQ1:} How do reported repository-repair outcomes differ with evolution enabled? \textbf{RQ2:} How do versioned capability changes enter subsequent application work, and when do they fall short of task success? SWE-bench Pro \citep{deng2025swebenchproaiagents} supplies the coding task setting; the reported comparison provides aggregate task outcomes. Six historical cases provide finer-grained evidence from archived versions, calls, comparisons, and product revisions. We analyze these units separately: a component call, a connected development sequence, and a benchmark task carry different forms of evidence.}

\subsection{SWE-bench Pro Public: task-level comparison}
\para{eval:settings}{\textbf{Comparison setup.} The current coding configuration pair uses the same model route, initial task interface, planning support, and resource limits. The evolution-enabled condition additionally mounts the adoption Tool and self-evolving Skill; the baseline omits them. Table~\ref{tab:swe-settings} summarizes this intended intervention. The declared roster contains one coordinator, so the coding comparison does not test the contribution of multi-agent coordination. Launch-time overrides must be recorded to establish the model and resources actually used for a run.}

\begin{table}[t]
\caption{Inspected SWE-bench Pro configuration pair. Both conditions share the entries in the first four rows. These are configuration defaults, not verified realized settings of the external run.}
\label{tab:swe-settings}
\small\begin{tabularx}{\linewidth}{@{}lY@{}}
\toprule Setting & Configuration \\
\midrule
Model and roster & Claude Opus 5 (configured route); one coordinator \\
Limits per task & 400 coordinator steps; 7,200 seconds; 3,000,000 tokens \\
Shared interface & Shell, completion, inspection, file access, planning, and memory \\
Grading boundary & Frozen patch submitted for final host-side evaluation \\
Evolution enabled & Adoption Tool and self-evolving Skill available \\
Evolution disabled & Same initial task interface, without those two capabilities \\
\bottomrule\end{tabularx}
\end{table}

\para{eval:protocol}{\textbf{Evaluation and available evidence.} The current launcher submits a frozen patch for host-side grading and withholds final grading feedback during solving. The reported totals come from a separate-machine execution. Its raw per-task results, realized configuration, and complete cost records were not available for this analysis; the configuration in Table~\ref{tab:swe-settings} therefore describes the inspected setup rather than a verified execution manifest for those totals. Appendix~\ref{app:protocol} details the grading boundary and records needed for reconciliation.}

\begin{figure}[t]
\centering
\includegraphics[width=\linewidth]{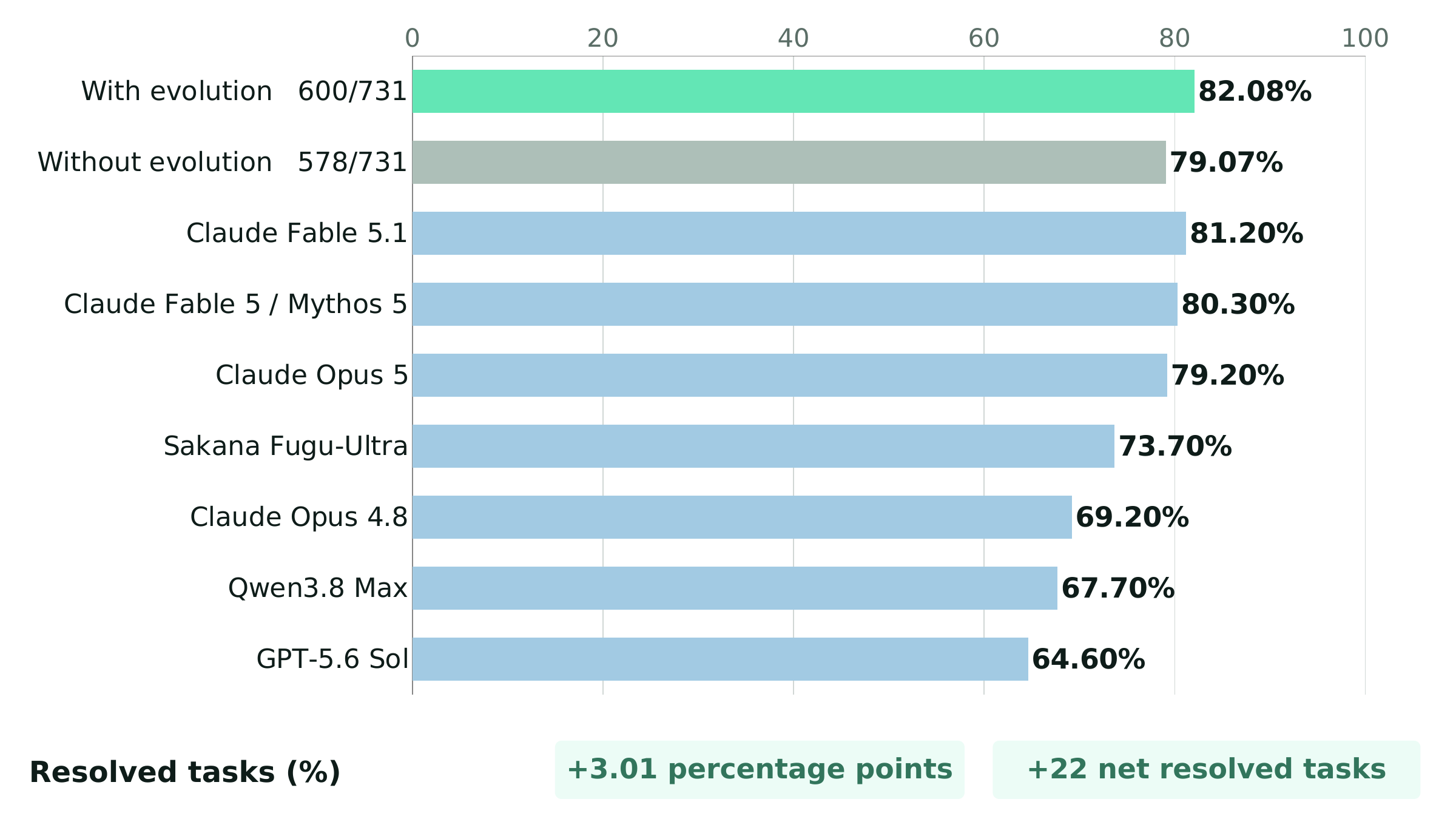}
\caption{\textbf{SWE-bench Pro Public results.} The top pair shows the team-reported comparison: 600/731 with evolution versus 578/731 without it, a gain of 3.01 percentage points and 22 net resolved tasks. The remaining bars reproduce the website's 19 September 2026 reference snapshot. Models, harnesses, budgets, and grading settings differ across external reports; the common zero-based axis does not imply a controlled comparison.}
\label{fig:swe}
\end{figure}

\para{eval:score}{\textbf{Results.} The team reports 600/731 resolved tasks with evolution and 578/731 without evolution. The corresponding rates are 82.08\% and 79.07\%, an absolute difference of 3.01 percentage points and a net increase of 22 resolved tasks. The top pair in Figure~\ref{fig:swe} presents this internal comparison. The reported comparison therefore favors the evolution-enabled condition.}

\para{eval:external}{The external rows provide context from the website's frozen 19 September 2026 snapshot, which cites the CodingFleet compilation updated on 11 September 2026.\footnote{Website data: \href{https://dvampire.github.io/AgentEvolver/assets/demos/benchmark.json}{AgentEvolver benchmark snapshot}; compilation: \href{https://codingfleet.com/blog/swe-bench-pro-leaderboard-2026/}{CodingFleet SWE-bench Pro results}.} Their original model and system labels are preserved. Differences in models, harnesses, budgets, and grading conditions prevent these rows from isolating an AgentEvolver effect or establishing a controlled product ranking. The relevant internal contrast for RQ1 is the reported with-evolution versus without-evolution pair.}

\para{eval:interpretation}{\textbf{Interpretation.} A net increase of 22 does not reveal how many tasks improved or regressed. If $b$ counts tasks solved only with evolution and $c$ counts those solved only without it, the totals determine $b-c=22$ but not either count. Task identities must also be reconciled before assuming exact pairing. Consequently, the aggregates do not support a paired significance test, a gain-per-compute estimate, or attribution to any individual subsystem. The evidence for RQ1 is therefore the aggregate difference between the two reported conditions.}

\subsection{Six application cases: products and the capabilities behind them}
\begin{figure}[t]
\centering
\includegraphics[width=\linewidth]{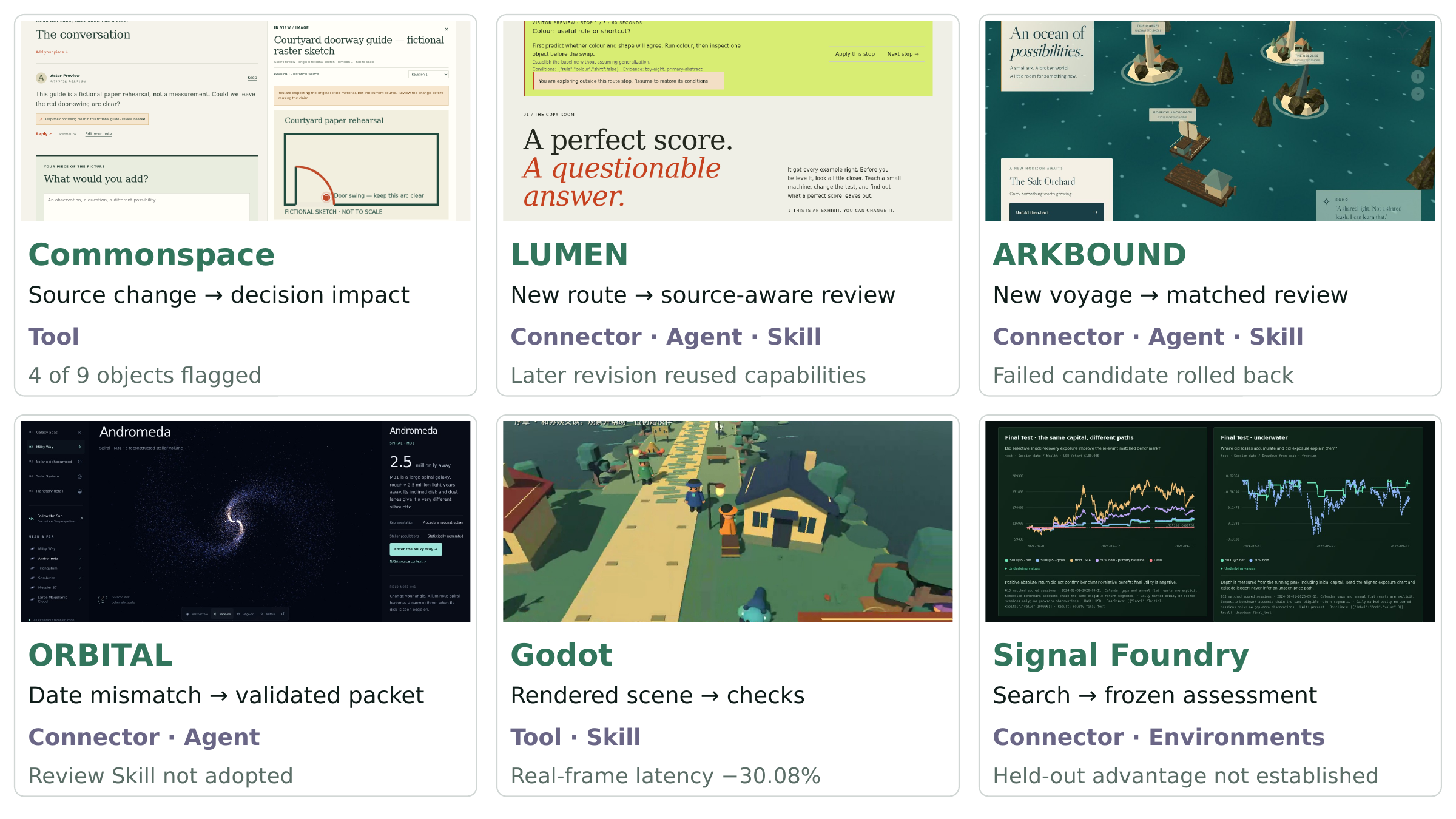}
\caption{\textbf{Six applications and their capability changes.} Original product views connect each task to the component families developed and an observed outcome. Images illustrate artifacts; component records and measurements support the claims. ORBITAL did not adopt the required review Skill, and Signal Foundry did not establish held-out strategy superiority. The cases are selected historical studies, not independent replications.}
\label{fig:cases}
\end{figure}

\para{eval:cases}{\textbf{Case selection.} We examine four websites, a connected Godot development sequence, and factor-strategy research selected from the historical demonstration collection. Selection was purposive: the available records expose different capability changes and allow the relation between local evaluation and final product state to be examined. The collection includes partial and unsuccessful outcomes but is not an exhaustive sample of all runs. Figure~\ref{fig:cases} presents the products; Table~\ref{tab:cases} connects their capability changes to measured observations and interpretive limits.}

\para{eval:caseprotocol}{\textbf{Analysis protocol.} The unit is a capability-change episode within a continuing project. For each episode, we reconstruct the task requirement, observed limitation, candidate version, comparison, retention decision, and later consumer, then examine the final task outcome separately. Executed-call observations support behavioral claims; builder judgments describe the local assessment; screenshots document visible product states. The case set establishes changes and subsequent use in five families: Tool, Skill, Agent, Connector, and Environment. Appendix~\ref{app:cases} records the case-specific evidence and the original visual sequences.}

\para{eval:applicationvision}{\textbf{Application design.} The four websites explore continued collaboration after an initial release. A user changes a source, requests a new interaction, or asks for a different explanation; the builder must revise the product while preserving earlier work and identifying assumptions that no longer hold. Commonspace applies this idea to community evidence, LUMEN to scientific inquiry, ARKBOUND to comparative play, and ORBITAL to astronomical source data. Recurrent needs motivate reusable interfaces, specialist behavior, and review methods. The recorded revisions instantiate this design through builder instructions and self-checks, without an independent-user longitudinal study.}

\begin{table}[t]
\caption{Six historical cases: component changes, observed effects, and evidence boundaries. Each count refers to the stated artifact or comparison fixture.}
\label{tab:cases}
\small
\begin{tabularx}{\linewidth}{@{}L{.17\linewidth}YY@{}}
\toprule
Case & Observed change or result & Interpretive boundary \\
\midrule
Commonspace & An impact-tracing Tool identifies 4 affected objects in a 9-object revision graph. & One connected work sequence; no population-level accuracy estimate. \\
LUMEN & Connector, Agent, and Skill retained with later use; a review fixture exposes 5 issues missed by structural validation. & Families prescribed by the task; reviewer findings are not a visitor study. \\
ARKBOUND & Asset Connector obtains 666,655 bytes within a 700,000-byte budget; an Agent candidate is rolled back and revised. & Budget compliance and local review do not measure player experience. \\
ORBITAL & Connector rejects date-misaligned data and is later consumed; specialist Agent is improved. & Required review Skill is not adopted; overall objective remains incomplete. \\
Godot & Brightness Tool median latency falls from 1,345.68 to 940.92 ms on one real frame; font Skill reaches a later consumer. & Three timings per arm; no game-FPS or full-game completion claim. \\
Signal Foundry & Research artifacts include 77 factors and 40 formal strategy versions; final selected strategy lacks the required advantage. & Search volume and positive raw return do not establish robust outperformance. \\
\bottomrule
\end{tabularx}
\end{table}

\subsubsection{Four websites as continuing design work}
\para{eval:commonstory}{Commonspace connects a source revision to the decisions that depend on it. The retained impact-tracing Tool identifies four affected objects among nine in the revision graph. A later consumer uses the operation, and the website exposes the consequences through a revision report and navigation to an affected action. The important capability is locating review work after the source changes, while preserving the history against which the earlier decision was made.}

\para{eval:lumenstory}{LUMEN makes the dependency between source, design, and review explicit. The Connector preserves source identity and metadata-only qualifications; the specialist authors an inquiry within those limits; the Skill checks its claims and treatment of historical observations. On a defect fixture that passes structural validation, the review method identifies five issues. All three retained versions enter later project work, showing how a revised source contract reaches the generated content and the method used to review it.}

\para{eval:arkstory}{ARKBOUND combines bounded asset acquisition with comparable play conditions. A Connector obtains 666,655 bytes within a 700,000-byte budget, and a paired-play Skill restores the starting state used for route comparisons. An Agent candidate is rolled back when the intended prompt change is absent from the running behavior, before a later version is retained. This failure identifies a concrete reason to evaluate the loaded component and verify that its revised instructions govern the observed task behavior.}

\para{eval:orbitalstory}{ORBITAL reveals a source error hidden by a plausible display: the response contains the expected number of states but covers the wrong dates. The revised Connector rejects that fixture and later retrieves an identified Mars packet. A specialist distinguishes source, model, and presentation changes. The required review Skill lacks adoption and use evidence, leaving the overall objective incomplete despite those retained local improvements.}

\subsubsection{Game development and research infrastructure}
\para{eval:gamevision}{Godot development requires inspecting realized engine behavior. A font can advertise a weight axis while the rendered face ignores the requested setting. The retained font-realization Skill probes the actual resource and rendered text, then reaches a later world-space label consumer. Its successful use is distinct from delivery of the unfinished game.}
\para{eval:godot}{A brightness Tool supplies a numerical component comparison. On one real-frame fixture, median latency falls from 1,345.68 to 940.92 milliseconds across three calls per implementation: a 30.08\% reduction with matched returned metrics. A later lighting review consumes the retained Tool on a newly rendered frame. Appendix~\ref{app:godot} separates this local operation measurement from the synthetic workload, subsequent use, and game performance.}

\para{eval:researchstory}{Signal Foundry improves the infrastructure for factor-strategy research. A Connector qualifies data, and two laboratory Environments materialize factors, evaluate strategies, and replay the selected three-factor S010@5 binding. The run contains 77 evaluated factor definitions and 40 formal strategy versions, but the selected strategy fails its comparative utility criterion. Successful replay establishes computational continuity; the selected strategy still lacks comparative advantage.}

\subsubsection{Cross-case findings and open questions}

\para{eval:synthesis}{Three patterns answer RQ2. \textbf{Semantic failures motivate useful revisions:} ORBITAL's source dates, Godot's realized font weight, and LUMEN's interpretation of historical observations expose gaps that syntax or schema validation misses. \textbf{The later consumer is the connection to task work:} impact tracing, route review, engine diagnosis, and frozen replay each enter an identified operation after their initial comparison. \textbf{Local improvement can coexist with an unmet objective:} ORBITAL lacks a required review component, Godot remains an unfinished game, and Signal Foundry's selected strategy fails its comparative criterion. The cases therefore support an analysis of specific capability changes and their use, rather than a success rate for the entire platform.}

\para{eval:open}{The cases also identify testable questions for further study. Prescribed component families leave open how an unconstrained agent selects a change surface. Connected-project reuse leaves open whether a frozen capability library helps independent tasks. A local operation speedup leaves open whether its savings recover development cost. Appendix~\ref{app:protocol} translates these gaps into matched comparisons, held-out consumer studies, and mechanism-specific ablations.}

\section{Discussion and Limitations}
\label{sec:discussion}
\para{discussion:judgment}{\textbf{Evaluation quality.} Versioned calls and observations make claims inspectable but do not validate the evaluator's fixtures, baselines, or judgments, especially when the same agent proposes and evaluates a revision. Repeated selection can also overfit development feedback \citep{JMLR:v11:cawley10a,dwork2015generalizationadaptivedataanalysis}. Stronger conclusions require independent acceptance tests, preserved failed attempts, and a frozen candidate before final evaluation. Feedback used for another revision becomes development evidence. Signal Foundry lacked enforced isolation of final evaluation data.}

\para{discussion:selection}{\textbf{Coverage, attribution, and cost.} The six selected cases, some with prescribed families and connected sessions, do not establish platform-wide success rates or gains across all eight families. Website checks also leave user satisfaction and learning outcomes unmeasured. The SWE comparison remains team reported; its raw run package and paired task outcomes are needed for attribution and uncertainty estimates. Generalization and cost effectiveness require frozen capabilities tested on independent consumers under matched budgets, accounting for regressions, abandoned candidates, and total development cost.}

\para{discussion:engineering}{\textbf{Execution and reproducibility.} Resource coordination relies on declared access, summaries can omit facts, and interrupted external operations can leave uncertain effects. Admission may provisionally activate a candidate before functional evaluation, while archived source can depend on changing packages or providers. Reproduction therefore requires inputs, dependency and environment records, and execution history alongside source bytes \citep{Sandve_2013}. Mechanism-specific ablations and recovery tests should assess how Plan, Runtime, and context management affect final task acceptance under these conditions.}

\section{Conclusion}
\label{sec:conclusion}
\para{conclusion:all}{AgentEvolver connects task execution to the development of reusable software capabilities. Its central design separates the deliverable, archived and active component versions, and the observations supporting retention. Eight evolution surfaces share execution, planning, and context mechanisms so that capability changes can participate in continuing work. The reported coding comparison favors evolution, and six application cases show how specific changes are evaluated and consumed, including failures to meet the final objective. Together, these observations motivate evaluating an agent through both what it delivers and what it retains. Establishing when retained capabilities generalize and repay their development cost is the next empirical step.}

\label{main:end}
\label{references:start}
\bibliographystyle{iclr2027_conference}
\bibliography{iclr2027_conference}
\label{references:end}
\clearpage
\appendix

\section{Core System Capabilities}
\label{app:core}
\label{app:contracts}
\begin{figure}[t]
\centering\includegraphics[width=\linewidth]{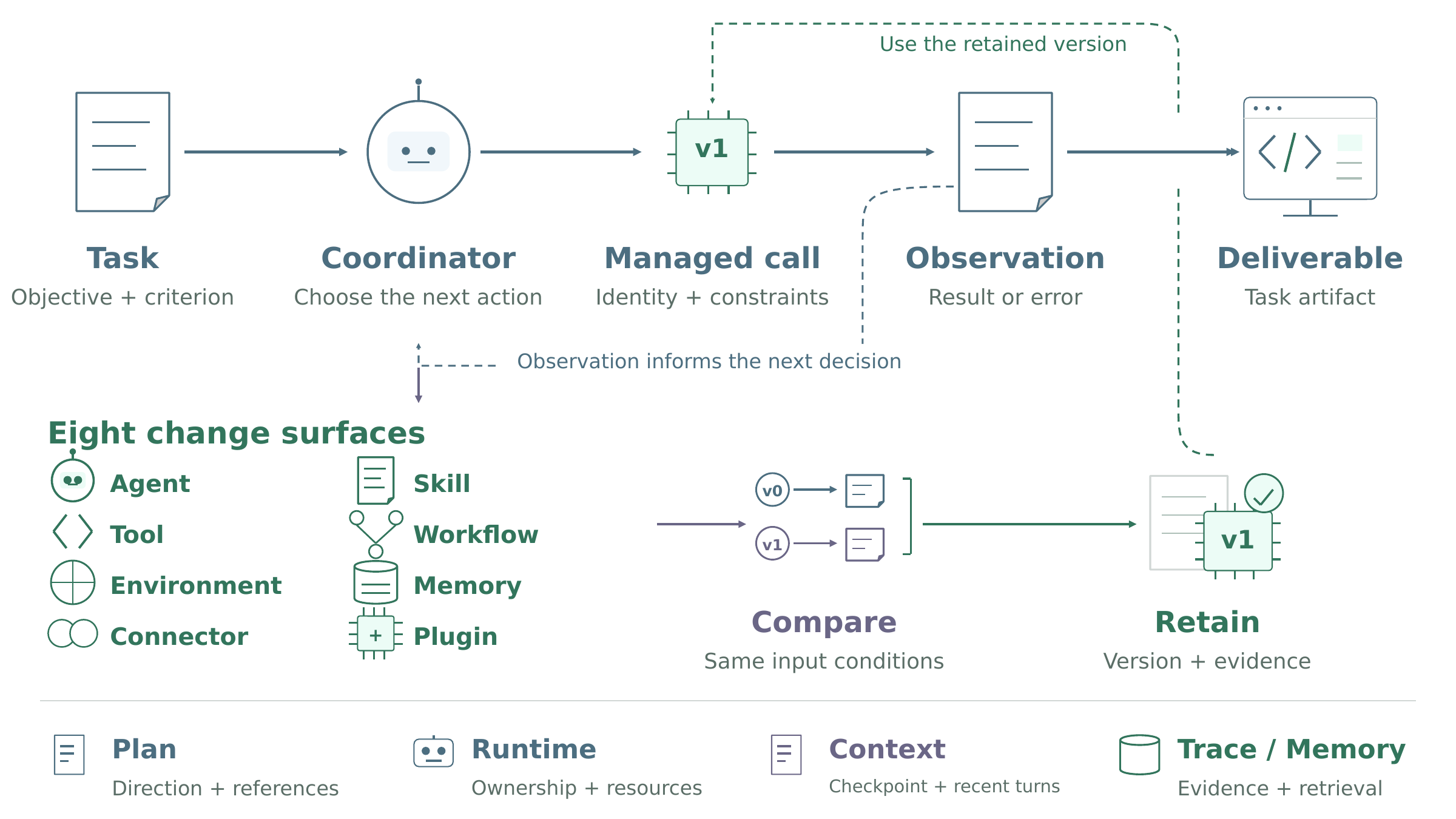}
\caption{\textbf{Task delivery and capability accumulation.} The execution path produces an artifact; an observed gap can also initiate a component revision and comparison. Retained versions return to later execution. Plan, Runtime, Context, and Trace/Memory provide complementary support. The diagram expresses functional relationships, not measured contributions from an ablation.}
\label{fig:app-system}
\end{figure}

\para{core:overview}{This appendix explains how AgentEvolver supports capability development during an unfinished task. The organization follows the life of the work: identifying what can change, assigning execution, preserving the objective, managing a growing interaction history, composing operations, and retaining evidence. Figure~\ref{fig:app-system} summarizes these relationships. The mechanisms are complementary. A versioned component needs an execution context; an execution needs direction; and a decision to retain a change needs observations that remain interpretable after the interaction has moved on.}

\subsection{Two coupled forms of progress}
\para{core:twooutputs}{An ordinary task state describes the artifact being produced and the conditions it must satisfy. An evolving system also maintains the means available for changing that artifact. A website may acquire a new inquiry route while its builder acquires a source-review method. A repository patch may repair a defect while its agent develops a diagnostic operation. These outcomes can reinforce one another, but their identities and acceptance criteria differ. A completed deliverable does not establish that a reusable component improved, and retaining a useful component does not establish task completion.}
\para{core:state}{Equation~\ref{eq:state} distinguishes the deliverable $D_t$, the archive $\mathcal{L}_t$, and the active version map $A_t$. Task actions use the active collection to change the deliverable. A capability revision can add an archived version and alter the active selection, while evaluation and decision records explain why it was retained. Structural admission may provisionally activate a candidate before that judgment. Keeping these states separate prevents availability from being interpreted as demonstrated improvement.}
\para{core:objective}{The original task supplies the reason to develop a capability. An expensive operation may motivate a Tool revision; a recurring interpretation error may motivate a Skill; incompatible source semantics may motivate a Connector. The relevant comparison follows from the intended consumer. Faster execution with changed outputs is different from faster execution with matched outputs, just as a stricter rejection rule is different from a higher task completion rate. The framework provides identities and evidence records for these comparisons under task-specific evaluation criteria.}
\para{core:opportunity}{This perspective makes capability development selective. Not every local difficulty warrants a reusable abstraction. A one-off value correction may belong only in the deliverable, whereas a repeated error in reading source revisions may justify a retained method. A useful opportunity therefore names a concrete limitation, an operation or method that could be reused, and a prospective consumer. The system can support that reasoning without prescribing a fixed improvement threshold or requiring a certain number of new components per task.}

\subsection{Eight families, distinct operational roles}
\para{core:taxonomy}{The eight component families share discovery, identity, and version management but retain different execution semantics. Tool, Skill, Agent, Workflow, Connector, and Plugin appear in the callable capability roster. Environment supplies mounted state and native actions. Memory supplies retained context through its owning subsystem rather than appearing as another operation for the model to select. This separation avoids treating every extension as a stateless function while still making its implementation an explicit object of change.}
\begin{table}[t]
\caption{Evolution targets and the corresponding unit of evidence. A family identifies a change surface, not a guarantee that a particular task will benefit from changing it.}
\label{tab:app-families}
\small
\begin{tabularx}{\linewidth}{@{}lYY@{}}
\toprule
Family & What changes & What a useful comparison inspects \\
\midrule
Tool & Executable operation & Outputs, failures and cost on matched inputs \\
Skill & Operating method and supporting material & Application of the method to a real problem \\
Agent & Instructions, configuration and behavior & Loaded role, assignments and resulting behavior \\
Workflow & Reusable control structure & Terminal runs, branches and continuation \\
Connector & Provider-facing contracts & Source identity, semantics and rejection cases \\
Environment & State, observations and actions & Reset conditions, transitions and reproducibility \\
Memory & Retention or retrieval implementation & Retrieved evidence, omissions and ownership \\
Plugin & Shared service and related operations & Service state, operation behavior and cleanup \\
\bottomrule
\end{tabularx}
\end{table}
\para{core:tools}{A Tool packages a bounded operation with an input contract. Its output can be checked directly when the intended behavior is deterministic, or assessed under an explicit task criterion when interpretation remains necessary. A Skill instead describes how to work: which observations to obtain, which alternatives to compare, and which failure conditions to examine. Its value depends on application. Loading a Skill's instructions establishes access to the method, but evidence of use must include subsequent actions that follow that method.}
\para{core:agents}{An Agent definition specifies reusable behavior, including its instructions and configuration. Instantiating a second worker from an unchanged definition creates another execution, not another evolved entity. A prompt revision belongs to the Agent's behavior and version rather than to a ninth family. This distinction becomes important when a candidate appears to have improved on paper but the running worker still uses an earlier prompt. Evaluation must connect the admitted definition to the behavior actually invoked. ARKBOUND contains a concrete example of this discrepancy.}
\para{core:interfaces}{Connector and Plugin both support multiple related operations, but emphasize different responsibilities. A Connector provides a provider-facing contract: how a source is requested, normalized, qualified, and returned. A Plugin packages a service implementation and operations that may share configuration or a client. Their effects can depend on external state, so source bytes alone may not reproduce a result. A meaningful record also identifies the inputs, provider observations, and state assumptions under which the operation was evaluated.}
\para{core:environmentmemory}{Environment and Memory expose a further distinction between implementation and content. Updating the state of a research laboratory is ordinary use of an Environment; changing its materialization or evaluation behavior is a component change. Writing a useful note is a memory-content update; changing how notes are indexed or retrieved is evolution of the Memory implementation. Both forms of activity may improve task execution, but counting them together would obscure what the reusable system actually changed.}
\para{core:selection}{The family boundary helps select an appropriate intervention. A missing arithmetic operation can become a Tool, while a multi-step diagnostic procedure can remain a Skill that uses existing Tools. A specialist with its own context can become an Agent; recurring executable handoffs can become a Workflow. Several families may be needed when an error crosses interfaces. LUMEN's source, route-authoring, and review changes illustrate this composition, whereas the brightness measurement in Godot illustrates a narrowly scoped Tool change.}

\subsection{Versioned development and retention}
\para{core:identity}{A component is identified by its family, registered name, and version. Admission archives the payload associated with that identity and rejects different bytes submitted under an existing immutable version. The purpose is referential stability: an evaluation and a later consumer should refer to the same implementation rather than to a name whose meaning changes silently. Family-specific managers still own loading and domain state. A common identity does not erase differences between a method directory, an Agent definition, and a stateful service.}
\para{core:admit}{Structural admission checks whether a candidate can enter the system. An isolated interpreter checks the relevant loading, construction, and schema requirements before the admitted bytes are archived. This stage can provisionally activate the candidate, so structural acceptance precedes the final judgment of usefulness. Such activation makes evaluation possible but creates an obligation to inspect the active version afterward. A candidate that loads successfully may still have incorrect behavior, an unsuitable comparison, or an unfulfilled task-level purpose.}
\para{core:compare}{Functional evaluation identifies the candidate and baseline, states expected behavior, and associates observations with executed calls. A passing report requires nonempty passing cases, and the adoption mechanism checks the referenced observations and version consistency. These checks establish that the report concerns identifiable execution. They do not independently determine whether the selected fixtures are representative or whether the evaluator's interpretation is sound. Preserving observations separately from judgment allows those scientific questions to be revisited.}
\para{core:retain}{Retention connects a passing evaluation to the currently active archived version and records the decision in history. Rejection, replacement, rollback, and unload remain meaningful alternatives. A recorded intention to roll back is not itself the state change: the corresponding operation must execute, and the resulting active version must be checked. Keeping this distinction explicit is useful when a candidate fails after admission or when several revisions are considered in succession. The failed branch remains evidence about the development process.}
\para{core:consumer}{Subsequent use provides the link from a retained component back to ongoing work. A later consumer should identify the retained version, the new input or task context, and the observed result. Repeating the original fixture confirms repeatability under that fixture; applying the component to another step in the same project establishes task-local use. Independent-task transfer requires a separate evaluation design with an unexposed consumer. These levels form an evidence hierarchy, not interchangeable labels for successful registration.}
\para{core:dependencies}{Cross-family changes require compatibility as well as local improvement. A stricter source packet can invalidate an author's previous assumptions, and a revised authoring procedure can require a different reviewer. The useful unit of comparison may therefore be a chain of versions rather than one isolated component. The current architecture makes the versions addressable and their calls observable. The application cases provide examples of connected use, while systematic tests of compositional gains and regressions remain a separate empirical question.}

\subsection{Runtime ownership and process lifetimes}
\label{app:runtime}
\begin{figure}[t]
\centering\includegraphics[width=\linewidth]{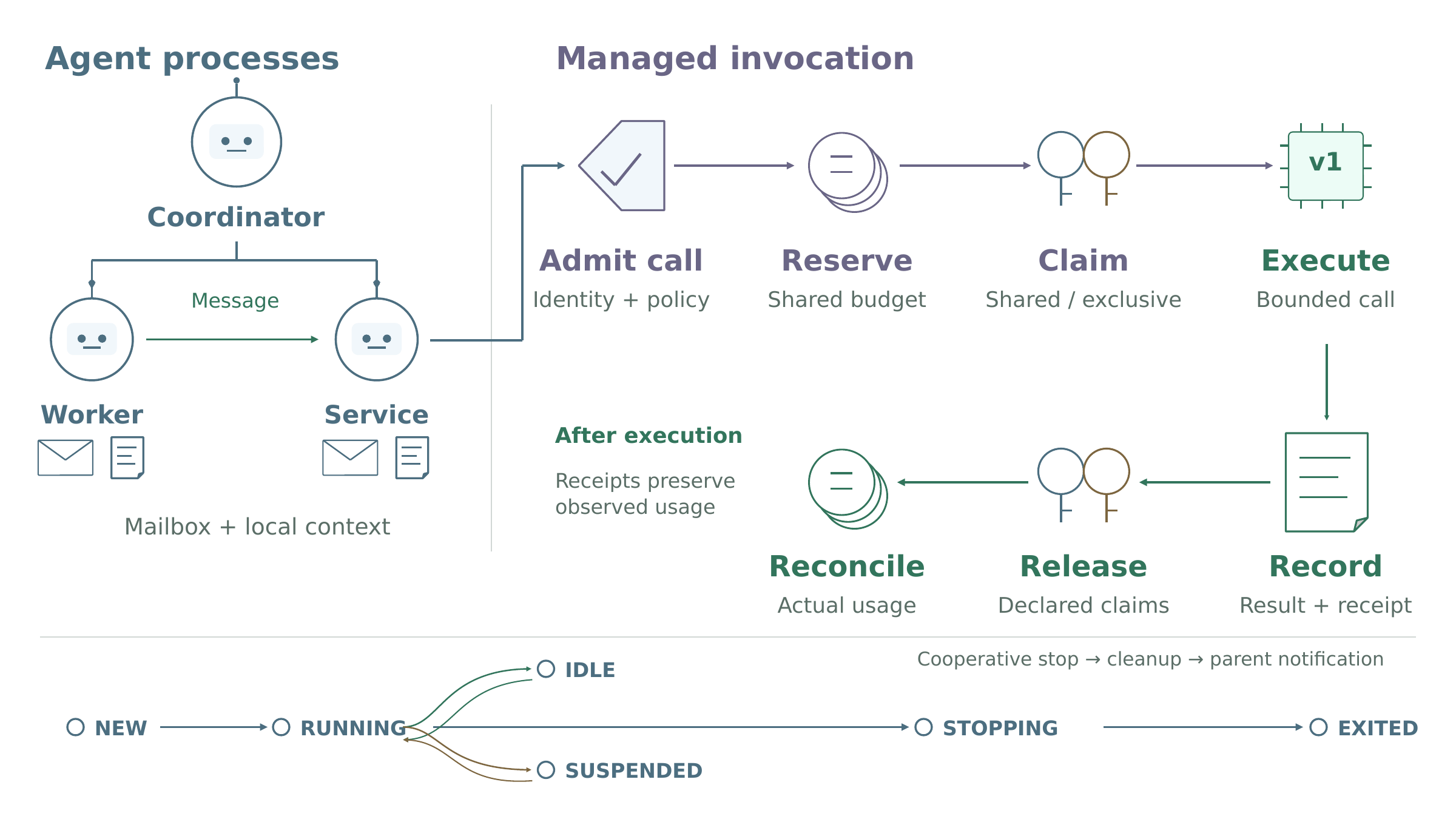}
\caption{\textbf{Runtime ownership, admission, and lifecycle.} A coordinator owns child Agent processes, each with its own context and mailbox. Bounded component calls instead pass through identity, policy, budget, and resource checks before execution and release. The lower state diagram concerns Agent lifetimes. Declared resource coordination does not imply distributed locking or automatic knowledge of external side effects.}
\label{fig:app-runtime}
\end{figure}

\para{runtime:reason}{Capability development takes place alongside product work and can outlive an individual model response. Runtime provides explicit ownership for these activities. An Agent process has an identity, an owning parent, a mailbox, and a lifecycle. A bounded component invocation has its own call identity and owner but does not allocate another Agent process. This distinction allows a coordinator to delegate reasoning to a specialist without turning every Tool call into a separately prompted participant. Figure~\ref{fig:app-runtime} shows the resulting organization.}
\para{runtime:modes}{Endpoint roles determine how work enters a process. A responder serves one request and exits. A service remains available for later addressed requests, while a subscriber receives work through registered topics. Resident processes preserve their own conversation across assignments. Failure of one resident assignment is recorded separately from the lifetime of the endpoint, so a failed turn need not imply that the service has disappeared. These roles share the same ownership and control mechanisms rather than separate orchestration systems.}
\para{runtime:states}{The process lifecycle distinguishes NEW, RUNNING, IDLE, SUSPENDED, STOPPING, and EXITED. IDLE means that a live endpoint is waiting for work; EXITED means its lifetime has ended. SUSPENDED preserves a process while preventing further progress at its control boundary. All endings pass through STOPPING so that cleanup and notification have one defined place. The distinction prevents a waiting specialist from being collected as finished and prevents task completion from hiding work still owned by the process.}
\para{runtime:messages}{Task messages, replies, reports, and events carry ordinary communication. Control requests govern suspension and stopping. Control is applied at safe points, such as the boundary between model steps or an explicit wait for messages. Consequently, a pause request does not promise instantaneous interruption of an arbitrary external operation. The semantics are cooperative where execution remains inside the Agent loop with stronger interruption depending on managed execution boundaries.}
\para{runtime:cleanup}{Ownership persists through cleanup. A parent stops and joins its children before reporting its own exit, and waiting for a process also waits for its cleanup path. If cleanup cannot finish before a shutdown deadline, the outstanding process identities and errors remain visible. This preserves an important distinction between a successfully answered task and resources that have actually been released. In evolving work, it also prevents a candidate-generation branch from silently continuing after its coordinator has declared completion.}
\para{runtime:example}{Consider a coordinator building a website while a specialist evaluates a source interface. The website can progress on independent layout work, but importing the revised source depends on the specialist's result. Runtime can represent the two activities and their ownership; the Plan records the dependency; the evaluation identifies the candidate version. Once the dependency is satisfied, the coordinator consumes the retained interface and checks the resulting product. This is a motivating execution pattern, not an additional historical experiment in the case dataset.}

\subsection{Resource coordination and bounded delegation}
\para{runtime:invocations}{Component invocations carry an owner, a parent call, timing, state, and declared resource claims. The invocation service coordinates admission across component families while their managers retain responsibility for their own state. Calls are lightweight coroutines, so this service does not imply CPU-level parallel execution. Its role is to decide when declared operations may coexist and to preserve their relationships during completion, failure, and cleanup.}
\para{runtime:claims}{Claims can be shared or exclusive. Shared claims on the same resource may coexist; an exclusive claim conflicts with other access to that resource. Path claims also account for overlap between ancestor and descendant paths, which matters when one operation owns a directory and another targets a file within it. Operations without sufficient independence information retain conservative sequencing. This mechanism is useful for coordinating known access within the runtime, but it cannot infer undeclared writes or enforce a distributed lock across arbitrary external services.}
\para{runtime:cancel}{Cancellation preserves the relation between an operation and its claimed resources. If a synchronous operation runs in a thread, requesting cancellation does not stop that thread immediately. The system waits for the operation before releasing its claims, avoiding the appearance that the resource is free while the underlying work continues. Externally executed effects introduce a second question: whether the effect happened before interruption. That question requires observation or reconciliation rather than being answered by the local cancellation flag.}
\para{runtime:brief}{Delegation uses a bounded assignment with a concise task, relevant files, declared read and write sets, and acceptance conditions. The child receives an independent context unless an explicit history fork is requested. Capability allowlists bound what is exposed to that child; an empty allowlist grants no members of the corresponding family. Declared file sets document intended scope, but are not by themselves filesystem enforcement. The distinction matters when interpreting a successful delegation as evidence of either coordination or isolation.}
\para{runtime:worktree}{For supported one-shot coding assignments, an isolated worktree can separate the child's edits from its parent's working tree. The runtime owns that worktree, preserves a patch before cleanup, and returns the patch without automatically merging it. If preservation or cleanup fails, the location and error remain available for inspection. This facility supports reviewable concurrent development, within its supported one-shot coding scope.}
\para{runtime:fork}{Explicit history sharing passes portable task context and complete interaction groups rather than implicitly granting a child all of its parent's state. A parent's own permission to receive history does not automatically pass to further descendants. Forked evidence and a resumed native conversation are also different operations: provider-specific opaque state stays with its original conversation. A bounded reference supplies portable evidence rather than native continuation state.}

\subsection{Budgeting across the complete development process}
\para{budget:shared}{A root run and its descendants share resource accounting, including later turns of resident processes. Candidate construction, evaluation, failed attempts, and subsequent consumption therefore draw on the task's common budget. The root objective remains the relevant unit of cost. Reporting only the retained component's final invocation would omit the work required to obtain it and could make an expensive adaptation appear artificially efficient.}
\para{budget:reservation}{The model layer reserves an estimated input cost and output allowance before framework-visible requests. Shared reservations prevent concurrent children from independently spending the same remaining capacity. Usage receipts then reconcile the reservation with observed usage, and receipt identities prevent duplicate charging when both request-level and turn-level records refer to the same call. The controls depend on observable requests and reported provider usage.}
\para{budget:unknown}{Missing usage remains unknown rather than becoming zero. An interrupted or cancelled request can leave an unreconciled reservation whose estimate continues to count against the budget. Remaining capacity can still support other work, but the unresolved reservation is retained across explicit resume until it is reconciled with evidence. This conservative treatment preserves uncertainty that would otherwise disappear from an apparent cost total. It also separates measured tokens from estimated exposure and monetary prices that may not be available.}
\para{budget:comparison}{A fair comparison between an evolving and fixed system should match the allowed resources and report realized consumption. Equal upper limits alone do not establish equal effort. The evolving arm may spend part of its budget developing a component, while the fixed arm may use that capacity for another task attempt. A useful evaluation reports total tokens, time, failed requests, and unresolved usage alongside accepted tasks. The current coding aggregates do not supply this full paired accounting, so this report does not infer cost efficiency from their accuracy difference.}

\subsection{Plan: persistent direction for changing work}
\label{app:planning}
\para{plan:purpose}{The Plan represents the coordinator's evolving understanding of the task. It records what remains to be accomplished, which dependencies are unresolved, and where the evidence for completed work can be found. It is neither another model nor an execution engine. This separation permits the coordinator to revise its approach as observations change while keeping the task objective explicit. Runtime remains responsible for whether work is running or finished; acceptance evidence remains responsible for whether the result satisfies its criterion.}
\begin{figure}[t]
\centering\includegraphics[width=\linewidth]{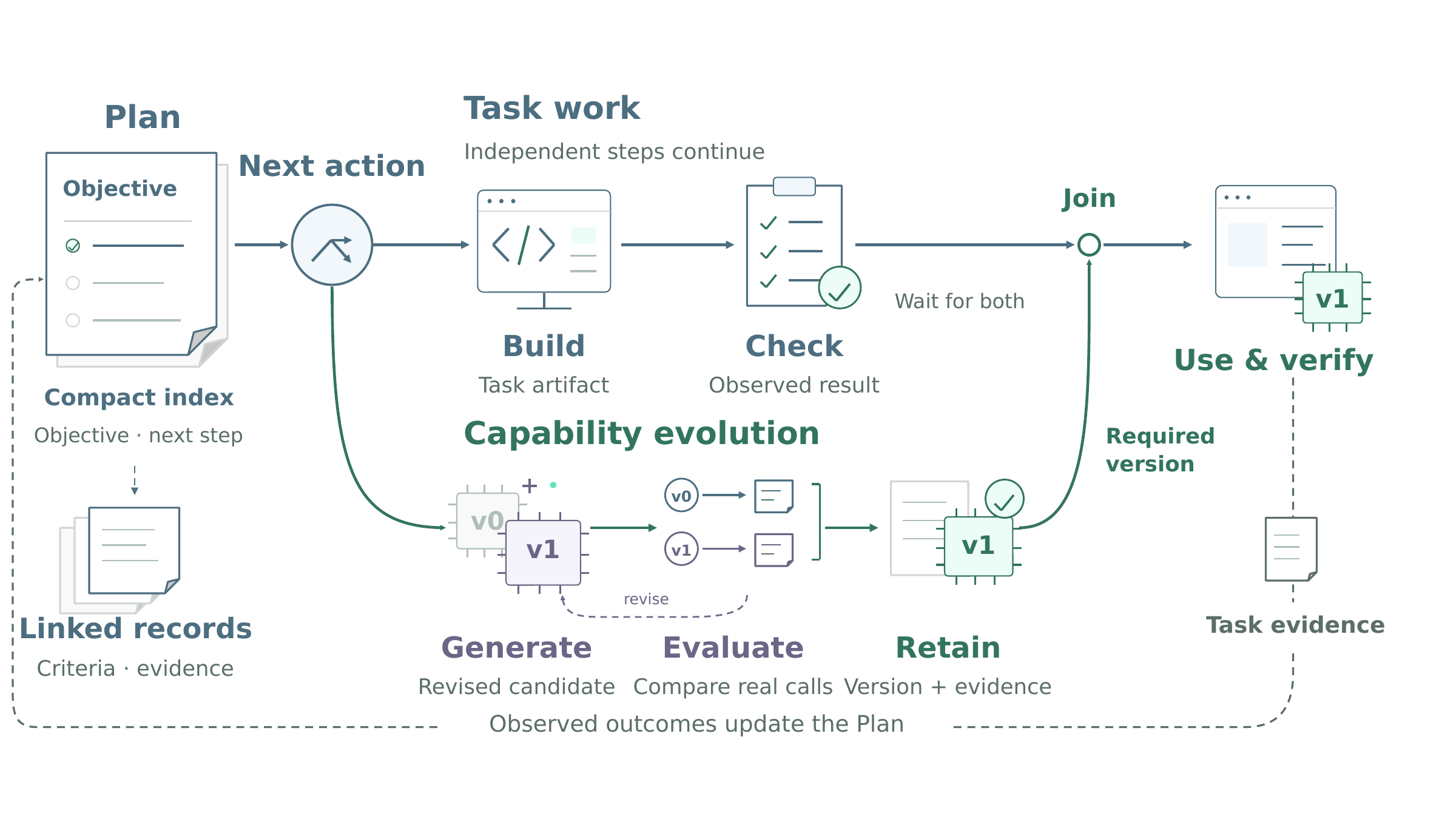}
\caption{\textbf{Persistent planning and dependency-aware execution.} A compact index links to detailed records and informs the next model request. Product work and qualified capability development may proceed concurrently; dependent consumers wait for the needed results. Outcomes feed back into the authored plan. Plan status, Runtime state, and acceptance evidence remain distinct.}
\label{fig:app-plan}
\end{figure}
\para{plan:index}{The mechanism separates a compact index from detailed working documents. The index is limited to 2,000 characters and identifies the current objective, progress, blockers, next action, and links to relevant records. Detailed reasoning, acceptance conditions, comparisons, and decisions remain in the plan documents. The agent authors these records with ordinary file operations rather than relying on an additional automatic summarizer. Changes to the index should follow meaningful progress so that its direction and supporting documents stay consistent.}
\para{plan:records}{This design manages attention through references. A model request receives a concise account of current work; a decision that depends on an earlier observation can explicitly retrieve the detailed record. The index may omit detail, but it should preserve the route to that detail. If an index is too long, truncation is visible and the original location remains available. Missing or unreadable records are reported as such instead of being filled with inferred progress.}
\para{plan:ownership}{Coordinators own planning and replanning; leaf workers receive bounded assignments and return evidence. A specialist does not automatically become a second owner of the full project plan. In coding tasks, plan records remain outside the submitted business repository so that planning activity does not contaminate the deliverable patch. Changing the workspace or using a child worktree does not redefine the coordinator's plan as part of that child's output.}
\para{plan:opportunities}{When planning and the configured evolution policy are enabled, the detailed plan tracks explicit evolution opportunities. An opportunity records the limitation that motivates it, the reusable operation or method, an intended consumer, expected benefit, relevant capability inspection, and a small comparison or reuse experiment. Its status includes the reason for proceeding or deferring. The record links proposed capability work to a concrete task need.}
\para{plan:branches}{The configured policy asks qualified opportunities to proceed as background branches when independent product work remains available. Within such a branch, generation, evaluation, and adoption retain their dependency order. Product work that requires the candidate waits for the result, and the coordinator collects outstanding branches before completing. A concrete prerequisite, conflicting write, permission constraint, or resource limit can justify deferral. These are coordination rules for the agent's procedure, not evidence that background execution always improves performance.}
\para{plan:modes}{Planning support and approval are separate controls. Auto mode supports plan maintenance while allowing authorized execution. Plan mode gates mutating actions until an explicit exit from the review phase. Off mode disables automatic planning support. The effect gate also applies to programmatically composed calls; describing an operation in a plan does not grant permission to execute it. The resulting separation keeps the written approach, the authorization state, and the observed outcome from collapsing into one status label.}
\para{plan:reconcile}{After an interruption, the next step may be reconciliation rather than continuation. A plan can say that a source packet was saved even if the corresponding external operation was interrupted before confirmation. The coordinator should inspect the saved artifact or execution record before repeating the action or marking the dependency resolved. The Plan's contribution is to preserve the intended direction and relevant references; it does not independently certify historical claims written into it.}

\subsection{Context assembly and recoverable compression}
\label{app:compression}
\para{context:problem}{Long tasks accumulate source material, failed attempts, and decisions. Even when this material fits, a longer context does not ensure reliable retrieval of information at every position \citep{liu2023lostmiddlelanguagemodels}. Retaining everything indefinitely increases request size, while discarding history can erase the reason for a constraint or the observation behind a comparison. AgentEvolver addresses this tension by separating stable task anchors from replaceable interaction history, preserving recent complete exchanges, and maintaining references to archived material. Compression manages resources while retaining explicit failure and recovery behavior.}
\para{context:assembly}{The context envelope distinguishes fixed instructions and task material, a checkpoint, recent interactions, and an optional transient layer. In the current running loop, changed runtime observations and environment images are first recorded in the conversation; the loop does not supply a separate transient layer. The assembler retains that optional interface for other views. This matters for describing the Plan correctly: the current index is refreshed before a request, but its recorded observations can subsequently become part of the history eligible for compression.}
\para{context:observations}{Runtime observations are marked as framework observations rather than user feedback or approval. Changed observations enter the history, and repeated unchanged content is not presented as a newly authored user instruction. Environment images are likewise tied to available observations rather than obtained by silently polling during request assembly. These conventions preserve the meaning of what the model sees. They also give an inspection record a basis for distinguishing a newly observed state from a previous image or an agent's own description.}
\para{context:protocol}{Compaction operates at closed interaction boundaries. Recent assistant calls and their results remain together, avoiding orphan results or requests whose calls have disappeared. The envelope validates call identities and layer consistency before serialization. Keeping complete recent turns is particularly useful when the next decision depends on the exact parameters of a recent failure. Summaries may omit parameters or qualifications preserved in the original exchange.}
\begin{figure}[t]
\centering\includegraphics[width=\linewidth]{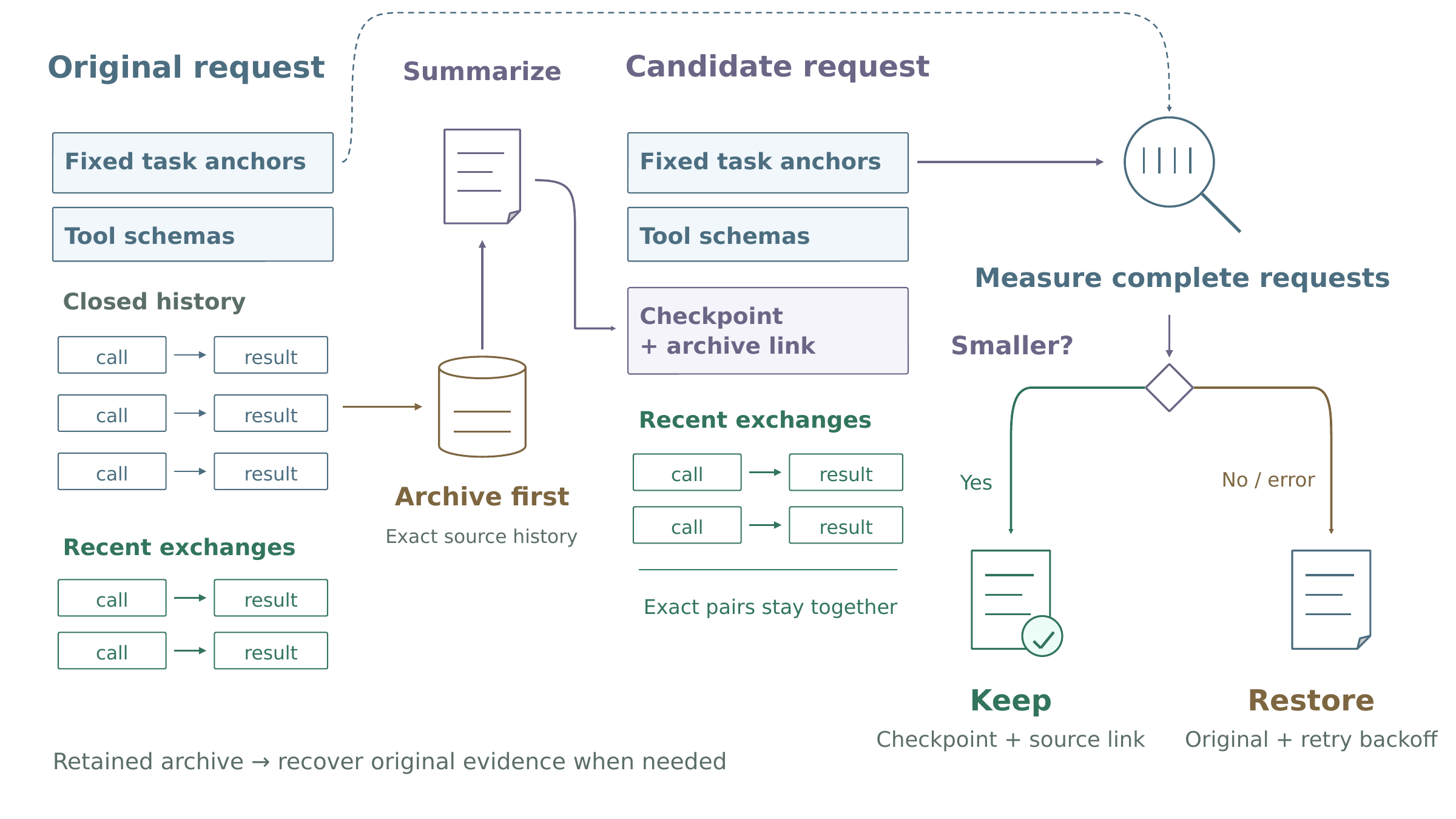}
\caption{\textbf{Context compression with a full-request savings check.} Both requests include fixed instructions and tool schemas. Thread-bound history is archived before summarization; recent complete turns also remain intact in the next request. A candidate is retained only when the complete request shrinks; otherwise the original history is restored with retry backoff. Layer sizes are schematic and do not represent measured token counts.}
\label{fig:app-context}
\end{figure}
\para{context:measure}{The pressure decision measures the serialized model request, including fixed instructions, checkpoint content, recent observations, and tool schemas. Measuring only the removable conversation body would miss a large stable prefix or an unexpectedly large native checkpoint. The implementation can calibrate the local estimate with observed provider input usage. Its policy therefore evaluates the actual request boundary at which context pressure arises rather than assuming that fewer visible history characters imply a smaller request.}
\para{context:defaults}{Current defaults retain four recent turns, use a full-input trigger of 100,000 tokens, and consider model-window pressure at a fraction of 0.85. The summary generation target is 2,048 tokens. These are configurable operating defaults, not empirically established optima for every model or task. Retention can be reduced toward one complete recent turn when necessary to seek headroom. The proposed recovery study in Appendix~\ref{app:protocol} would evaluate the tradeoff between preserved detail, request cost, and successful continuation.}
\para{context:archive}{For a conversation bound to a persistent thread, a fold first archives the complete closed conversation, including the previous checkpoint and any opaque provider state. If this archive cannot be written, the original history remains intact and the summarization call is not made. A successful checkpoint retains a locator for that archive, and subsequent archives preserve earlier locators. Thus repeated compression need not remove the route to old evidence, although deleting the containing session would still remove its stored artifacts. Unbound in-memory calls lack this archive guarantee.}
\para{context:candidate}{The summarization step produces a candidate checkpoint while preserving the fixed task anchors and selected recent complete interactions. Native checkpoints must have a readable companion before old history is replaced. A provider-specific checkpoint and its readable account serve different purposes: one supports the provider's continuation format, while the other permits human inspection and portable reference. Opaque state is not converted automatically between model routes, and explicit resume rejects an incompatible route instead of treating it as interchangeable text.}
\para{context:accept}{After constructing the candidate, the loop measures the full request again. If the native representation is less useful for request size or headroom, the already-produced readable companion can be considered without another model call. The smaller valid representation can be selected. If the candidate does not reduce the complete request, the original conversation is restored. This is a size-based acceptance condition for history replacement; it is separate from the soft output-length target used when generating the summary.}
\para{context:headroom}{The policy seeks headroom below the active limit rather than repeatedly operating at the boundary. A productive reduction with sufficient headroom resets the scheduled retry behavior. Unproductive or inadequate attempts trigger backoff, while actual capacity overflow can request recovery immediately. The implementation also limits repeated unproductive folds. These controls avoid spending another summarization attempt on essentially unchanged pressure at every step, while allowing a long productive run to compact repeatedly as its history grows.}
\para{context:semantics}{Request savings do not establish semantic completeness. A compacted account can omit a qualification, merge two similar versions, or retain a conclusion without its supporting observation. AgentEvolver does not apply an independent model audit that proves each summary preserves all acceptance conditions. The fixed task anchor, recent exact exchanges, fresh Plan index, and source archive reduce different risks, but none makes compression lossless. A coordinator facing a consequential uncertainty should retrieve the relevant original evidence rather than rely solely on the summary.}
\para{context:example}{Consider a website revision whose checkpoint says that source validation passed but omits that only metadata was available. The Plan can identify the route under revision; the archived source packet resolves the missing qualification. Similarly, a game-development summary may retain a font setting while omitting the engine measurement that motivated it. These illustrative cases explain the distinct roles of a compact task index, a working summary, and recoverable evidence. Their value depends on whether the agent retrieves the relevant detail when the next decision requires it.}
\para{context:evaluation}{A compression evaluation should therefore measure both resource behavior and continuation quality. Resource measurements include full request size, summarization cost, retries, and recovered headroom. Continuation measurements include retained acceptance conditions, correct version references, recovery of a needed source, and avoidance of repeated completed effects. Comparing only summary length would miss the task-level consequence. Conversely, a successful final task without request accounting would not isolate whether the compression policy reduced cost.}

\subsection{Workflow and programmatic composition}
\label{app:workflow}
\begin{figure}[t]
\centering\includegraphics[width=\linewidth]{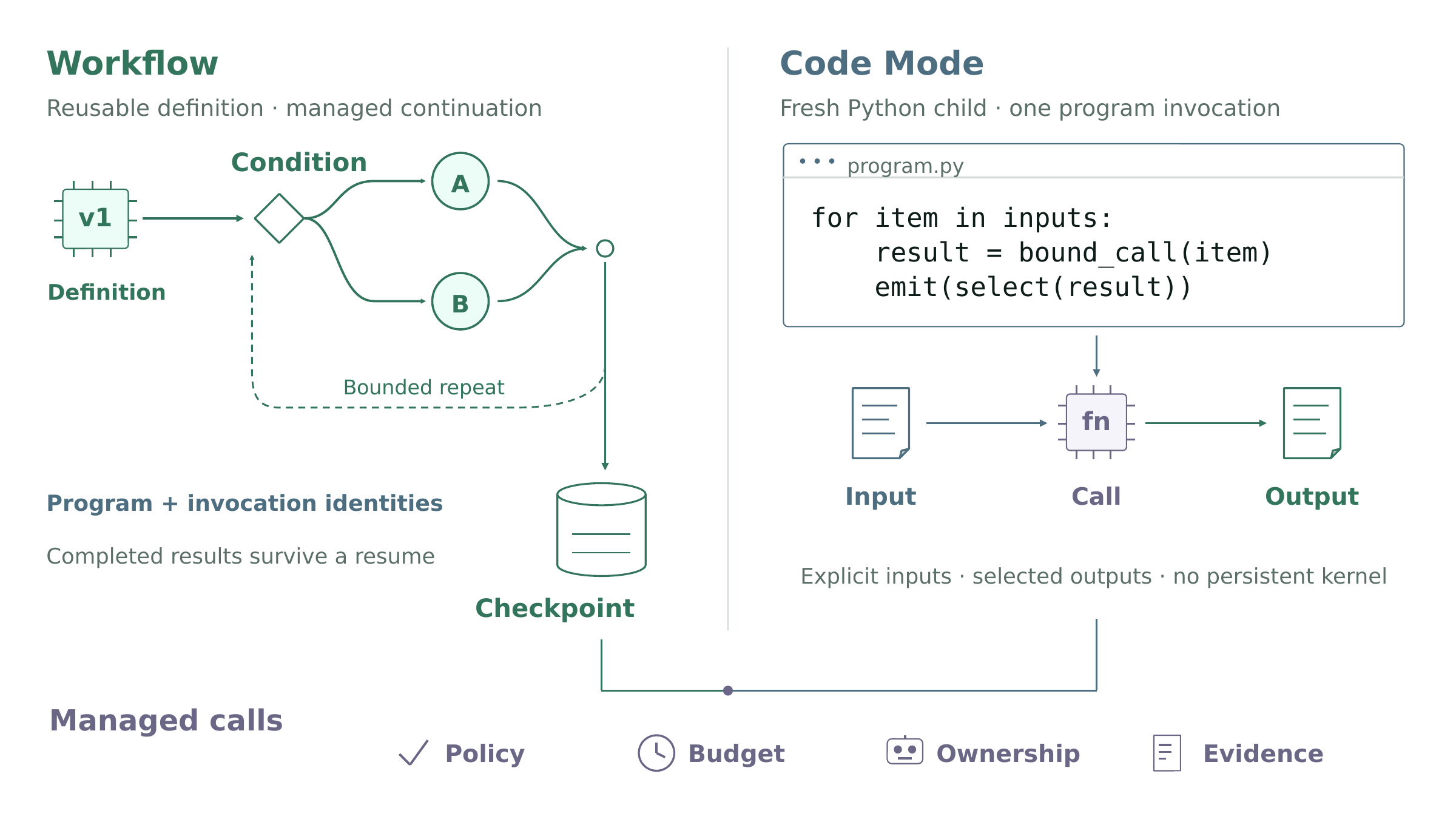}
\caption{\textbf{Reusable Workflow control and one-shot Code Mode.} A Workflow combines branches and bounded loops with checkpoints of completed execution. Code Mode starts a fresh Python child, composes bound operations, and returns selected outputs. The pseudocode illustrates composition rather than a public API. Both use managed calls; uncertain external effects need reconciliation.}
\label{fig:app-workflow}
\end{figure}

\para{workflow:distinction}{Plans, Workflows, and one-shot programs express different kinds of structure. A Plan is the coordinator's authored account of current work. A Workflow is a reusable executable control structure with a version and managed continuation. Code Mode is a programmatic interface for composing available operations within a single execution. Treating these as the same mechanism would obscure who chooses the next step, which state is retained, and what a later consumer can reuse. Figure~\ref{fig:app-workflow} contrasts the latter two.}
\para{workflow:language}{A Workflow is represented as a restricted HTML program that can be read and inspected before execution. The compiler admits defined control constructs and input contracts rather than executing arbitrary content embedded in a document. Branches, mappings, and bounded loops support dynamic control; the system is therefore more expressive than a single fixed DAG. Loops require explicit bounds, and limits constrain fan-out, nesting, concurrency, Agent count, and elapsed time. These controls bound the program's operation rather than determine whether its decisions are useful.}
\para{workflow:frames}{Execution separates the static definition from dynamic frames, concrete invocations, and individual attempts. A mapped item or loop round has its own frame; a frame can initiate an invocation; retrying that invocation produces attempts with distinct outcomes. This hierarchy allows the system to identify which part of a reusable procedure failed and which work completed before interruption. Retry attempts remain distinct from independently completed runs.}
\para{workflow:continuation}{Workflow state progresses through validation, readiness, execution, and verification to success or failure, with explicit pause and cancellation paths. Checkpoints preserve the executable program identity and completed progress. Resume can reuse completed invocation results while restarting unfinished work, and a changed program cannot silently reuse the same version's checkpoint. However, a checkpoint's local status does not settle whether an interrupted external side effect occurred. The ambiguous effect must be reconciled before retrying it.}
\para{workflow:versions}{The language schema, checkpoint representation, and Workflow behavior have separate versions. A behavior revision can change the procedure while retaining the same language format; a runtime checkpoint change can affect continuation without changing the task procedure. Keeping these identities distinct prevents compatibility claims from being inferred from a single version number. Evaluation of a Workflow should identify a real terminal run, its inputs, and its outcome rather than merely show that the source document passed parsing.}
\para{code:program}{Code Mode serves a different need: several related operations can be composed in one small program, with intermediate data processed before selected results return to the model. Each run uses a fresh Python child with bindings to available operations. It does not provide a persistent kernel whose hidden state automatically carries into the next program. This makes the program boundary explicit, while authorized external state remains accessible through ordinary operations.}
\para{code:control}{A bound call returns through the ordinary Tool pipeline, preserving policy checks, budget accounting, ownership, and relationships between parent and child calls. Code Mode does not provide an alternate privileged dispatcher. Failures can be handled as program exceptions, and partial output remains available when a program fails or exhausts its allotted execution. Keeping the nested call identities also matters for analysis: these calls were made inside the program and should not be misclassified as separate actions directly emitted by the model.}
\para{code:limits}{The fresh child-process boundary is not an operating-system security sandbox. Its permitted filesystem and service access depend on the surrounding execution environment. Similarly, selecting a smaller returned result can reduce material immediately placed in the model context, but does not establish a measured token or latency gain without comparison. The current report presents Code Mode as a controlled composition interface. Quantifying its efficiency would require matched operations, equivalent outputs, and complete request accounting.}
\para{workflow:example}{A source-review pipeline illustrates the division of labor. The Plan can state that a revised inquiry requires source verification and historical-consistency review. A reusable Workflow can encode the branch that requests a correction after a failed check. A one-shot program can read several packets, apply bound validators, and return a compact table for the next decision. The mechanisms support distinct levels of structure within the same evolving task.}

\subsection{Memory, conversation continuity, and historical retrieval}
\label{app:memory}
\para{memory:separation}{Prior agent architectures combine retained experience with reflection and planning \citep{park2023generativeagentsinteractivesimulacra,packer2024memgptllmsoperatingsystems}. Our system distinguishes an Agent's conversation from durable notes. A resident process appends new assignments and events to its own conversation, while a one-shot call starts fresh unless continuation is explicitly requested. Notes retain selected facts and references outside that immediate dialogue. Disabling automatic note injection does not erase conversation continuity. Conversation continuity therefore does not require copying every statement into notes.}
\para{memory:index}{The default notes interface provides a short index over files that the Agent can read when needed. The memory backend participates in indexing, so a revised Memory implementation can change how entries are ranked or exposed without replacing the Runtime's conversation. Note bodies are not all injected automatically. A compact index can omit complete entries with a notice and a route to the directory, preserving access to omitted material.}
\para{memory:ownership}{Notes are scoped to an actor within a session. Reopening that session can restore its notes, but another actor or session does not automatically acquire the same store. Changing the workspace does not relocate memory into the deliverable. These namespaces organize ownership; they are not themselves a security boundary against unrestricted host filesystem access. They also mean that persistent notes within a session should not be advertised as automatic, validated knowledge transfer across every future project.}
\para{memory:evolution}{An evolved Memory backend can alter retention or retrieval behavior. Demonstrating improvement would require showing which relevant records become available, which are omitted, and how the change affects a later consumer. The present cases do not establish such a controlled Memory improvement. The architectural support belongs in the system description, while the empirical coverage remains concentrated in Tool, Skill, Agent, Connector, and Environment changes. This distinction keeps the eight-family capability claim separate from the number of families actually evidenced in the selected cases.}
\para{memory:fallibility}{Notes are fallible references rather than instructions with higher authority. A stored observation should retain its origin and scope, especially when an earlier attempt failed. Repeating a model-written claim of success does not create independent evidence, and a note about a passing shell command does not establish that the task's acceptance conditions passed. Useful retention preserves the supporting observation and the conditions under which it was obtained, allowing later work to question an outdated or overly broad interpretation.}
\para{memory:history}{Historical retrieval complements the notes interface. Session-level reads can inspect an outline, search prior execution, or retrieve an exact bounded record from the authoritative history. Results report truncation and skipped malformed material rather than presenting a silently complete reconstruction. The route from a concise note or checkpoint to a specific original observation is particularly important when different component versions produced similar-looking results. Retrieval must recover the relevant identity as well as the text of the conclusion.}

\subsection{Trace, recovery, and evidence-preserving views}
\label{app:trace}
\begin{figure}[t]
\centering\includegraphics[width=\linewidth]{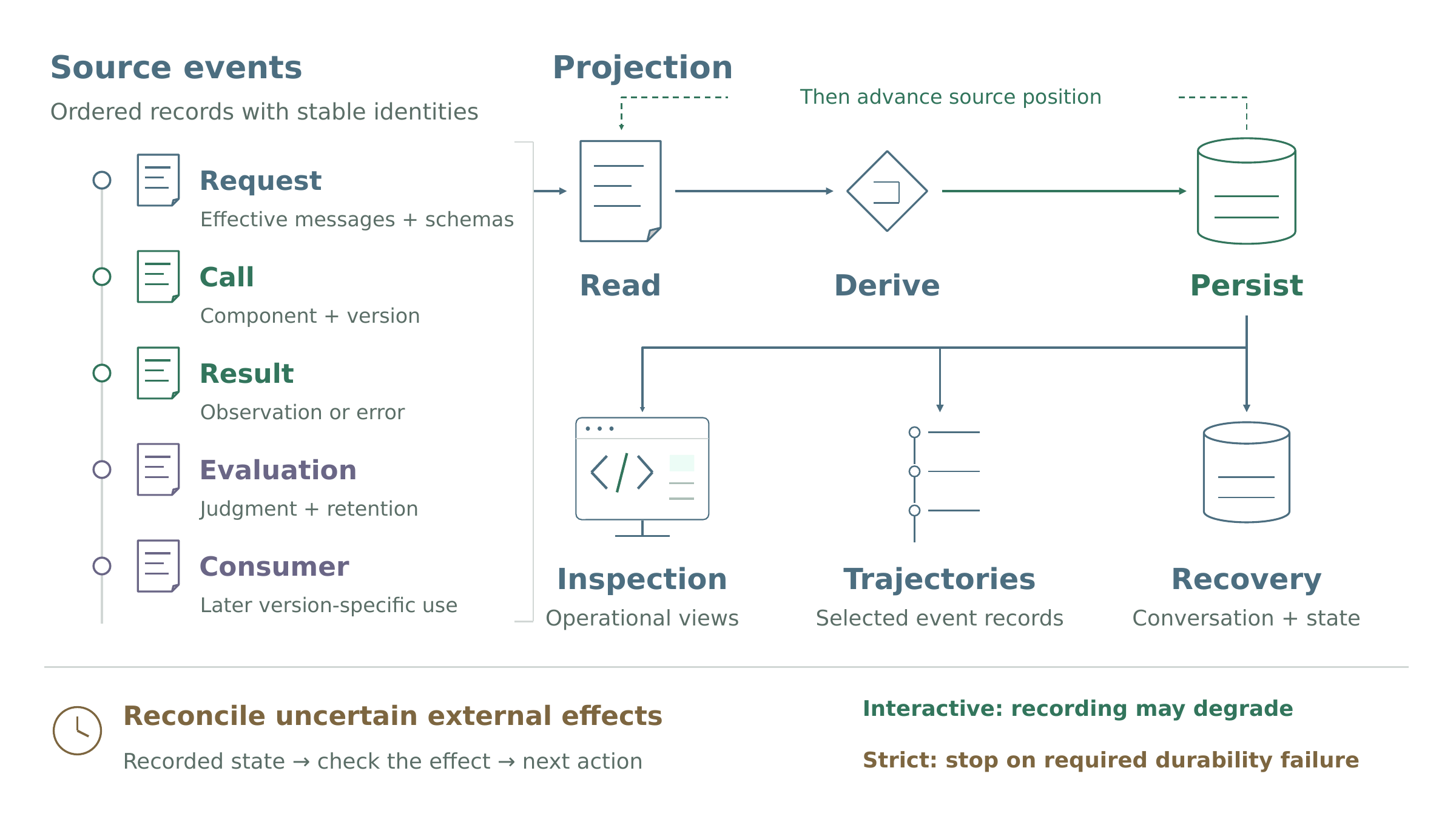}
\caption{\textbf{Evidence, derived views, and recovery.} Effective requests, execution observations, and retention decisions remain separate recorded events. Projections derive inspection views, trajectories, and recovery records, advancing their checkpoint after persisting output. Interactive tracing may degrade, whereas strict profiles stop when required durability fails. Recorded observations do not independently certify an authored judgment.}
\label{fig:app-trace}
\end{figure}

\para{trace:purpose}{Trace records the execution from which evaluations and retrospective analyses are derived. It connects effective model requests, actions, component results, and lifecycle events. A polished interface can show the resulting artifact, but these records establish which version was invoked and what observation was available when a decision was made. Figure~\ref{fig:app-trace} summarizes the relationship between source events and derived views. Traceability improves inspectability without turning a self-authored judgment into an independent acceptance result.}
\para{trace:request}{A model-request snapshot records the route actually attempted, effective behavior settings, messages, and complete tool schemas before provider dispatch. This is more informative than a configuration alias alone, because a fallback route or pressure-related excerpt can change what the model actually received. The snapshot has a content-derived identity so that later analyses can refer to the same request. When an observation was shortened for the request, the original result remains distinguishable from the excerpt that reached the model.}
\para{trace:results}{Tool results retain execution identity, component version, parent and root call relationships, outcome, timing, and stable error information. A policy-refused operation can still produce a paired result even though its body did not run. This is necessary to distinguish an unavailable operation from an executed failure. Parent-call relationships also preserve the structure of Code Mode and delegated execution, so analyses can distinguish nested work from direct model actions.}
\para{trace:integrity}{The default interactive profile treats Trace as observational: a recording problem can be reported while execution continues. Stricter training and high-risk profiles make required durability a precondition at meaningful boundaries, including before a provider request or a potentially mutating operation. A failure at such a boundary stops progression rather than pretending the missing evidence can be repaired by a model retry. Consequently, the durability guarantee must be stated with its selected profile; it should not be attributed to all runs by default.}
\para{trace:projection}{Derived views read committed events and preserve a position in the source history. A projection must durably store its output before advancing that position. After a crash, replaying an already applied suffix is preferable to skipping source events whose derived output was never saved. Monotonic positions and versioned projection implementations make this ordering inspectable. The resulting views include operational statistics and training-oriented trajectories, but this report does not claim that training was performed or that those views improved model weights.}
\para{trace:surface}{Compaction changes the active conversational view without requiring deletion of original events. A replacement identifies the source range it summarizes, while the append-only evidence remains available. The active history and the complete history therefore answer different questions: what context is being continued, and what events occurred. An inspection tool must preserve that distinction. Replaced observations should not be silently counted twice as new evidence, and a summary should not be treated as an independent second measurement of the underlying event.}
\para{recovery:continuation}{Recovery also separates kinds of state. A conversation snapshot can restore dialogue under the same bound owner, session, and model route. A mailbox record can preserve queued assignments and acknowledged turn completion. A browser checkpoint can preserve selected browser storage for a fresh page. None of these mechanisms serializes an arbitrary running process graph or proves that an external action completed. The restored record determines what is known locally; the uncertain effect determines what must be checked before execution proceeds.}
\para{recovery:uncertainty}{Possibly executed messages are not blindly replayed. If a process stopped after issuing an operation but before confirming its result, recovery must reconcile that operation using evidence about its external effect. Exactly-once behavior cannot be inferred from a local completed flag when the external service has no corresponding transaction guarantee. This limitation is especially relevant to evolving agents because a repeated registration, write, or evaluation could otherwise alter the version history that the resumed plan expects.}
\para{trace:privacy}{Request and trajectory records contain the material actually used during execution and therefore need the same handling as the original task data. Structured records avoid storing credentials as dedicated metadata fields, and endpoint fingerprints distinguish deployments without exposing complete endpoint addresses. This does not remove sensitive information that may legitimately appear inside task messages or observations. Exporting trajectories or sharing an evidence package requires selecting the relevant records and preserving the appropriate data boundary.}

\subsection{What the core mechanisms enable, and what remains to measure}
\para{core:synthesis}{The mechanisms form a connected account of execution. Component identities define what can change. Runtime assigns ownership and coordinates declared access. Plan preserves the task's direction and unresolved dependencies. Context management keeps a bounded working history with references to archived evidence. Workflow and Code Mode provide reusable and local forms of composition. Memory and Trace retain different kinds of information for later use and inspection. Together they support capability development without requiring the original objective to disappear into a separate optimization process.}
\para{core:scopeevidence}{Implementation support is broader than the measured evidence in this report. The six application cases show particular retained changes and consumers, while the coding comparison reports whole-system task outcomes. Neither isolates every core mechanism. Claims about recovery reliability, compression quality, parallel efficiency, and cross-task library value require targeted controls. The proposed studies specify those controls so that future results can be attributed to a mechanism rather than inferred from the existence of an interface or from a single successful application.}
\para{core:designlesson}{The central design principle is to preserve the connections needed to interpret progress: between a task and an opportunity, between a candidate and its executed version, between an observation and a decision, and between a retained component and a later consumer. Each connection can fail independently. Making them explicit permits useful partial outcomes to be retained while keeping incomplete objectives visible. This is the basis for studying global capability evolution as a process with observable artifacts and decisions, with distinct records of success and failure.}

\section{Case Studies: Evolving Products and Their Builders}
\label{app:cases}
\subsection{Study design and visual evidence}
\para{cases:design}{The six cases examine an application-oriented question: when a task changes, can the agent improve both the product and a reusable part of its way of working? Four website projects emphasize continuing design under revised sources, interactions, or interpretations. Godot development adds a native engine whose realized behavior must be inspected. Signal Foundry adds a research process in which reproducibility and strategy quality can diverge. The cases are selected historical development sequences, not a randomly sampled benchmark or six independent replications of one treatment.}
\para{cases:reconstruct}{Each reconstruction follows the same chain: application intent, observed limitation, candidate change, comparison, retention, subsequent consumer, and product outcome. Component archives and executed observations support the mechanism account. Product views show where the resulting capability enters the application. Session and version identities are retained in the evidence index so that a later review can locate the selected artifacts. The explanatory narrative groups related steps, but does not imply that every recorded operation occurred in a single uninterrupted run.}
\para{cases:frames}{Each case presents twelve real product views in a three-row, four-column grid. The website sequences use historical browser screenshots in recorded step order. The Godot sequence combines historical native screenshots with later recorded interaction. Signal Foundry follows the saved research report, including detail crops of its plots. These are selected views, not twelve independent experiments or a continuous real-time recording. Numbered reading guides identify what is visible; quantitative and adoption claims rely on the component observations discussed below.}
\para{cases:designintent}{The website studies model a continuing relationship between a user and a builder. Revised requirements can expose weaknesses in source handling, review, or specialist behavior that are not solved by changing layout alone. The intended outcome is a product that remains interpretable as it changes, together with methods and operations that support the next revision. Some tasks explicitly prescribe the families to evolve. Their evidence therefore concerns whether those changes were implemented and used, rather than whether an unconstrained agent independently chose the same design.}
\begin{table}[t]
\caption{Questions that organize the six case reconstructions. The right column identifies the application-level consequence to inspect separately from component adoption.}
\label{tab:case-design}
\small\begin{tabularx}{\linewidth}{@{}lYY@{}}
\toprule
Case & Capability question & Product question \\
\midrule
Commonspace & Can a revision be traced to dependent work? & Which existing conclusions need review? \\
LUMEN & Can source, design and review remain aligned? & Does a revised inquiry preserve evidence? \\
ARKBOUND & Can changed play be compared fairly? & Do routes and saved history remain coherent? \\
ORBITAL & Can a plausible but invalid packet be rejected? & Is the displayed interpretation qualified? \\
Godot & Can a method inspect realized engine behavior? & Did the intended visual change occur? \\
Signal Foundry & Can a frozen research binding be replayed? & Does the selected strategy meet its criterion? \\
\bottomrule\end{tabularx}
\end{table}

\subsection{Commonspace: Following a Revision into Its Consequences}
\label{app:commonspace}
\subsubsection{Application intent and observed gap}
\para{common:motivation}{Commonspace is a community collaboration website organized around materials, contributions, proposals, decisions, and follow-up actions. Its motivating design problem is continuity: a participant may return after a source changes, while earlier discussions and commitments remain visible. A useful builder must preserve the prior evidence and expose which conclusions need reconsideration. The historical sessions \code{93e9b80c} and \code{41857901} form a connected development sequence. They are examined as stages of one product's evolution rather than independent demonstrations.}
\begin{figure}[t]
\centering\includegraphics[width=\linewidth]{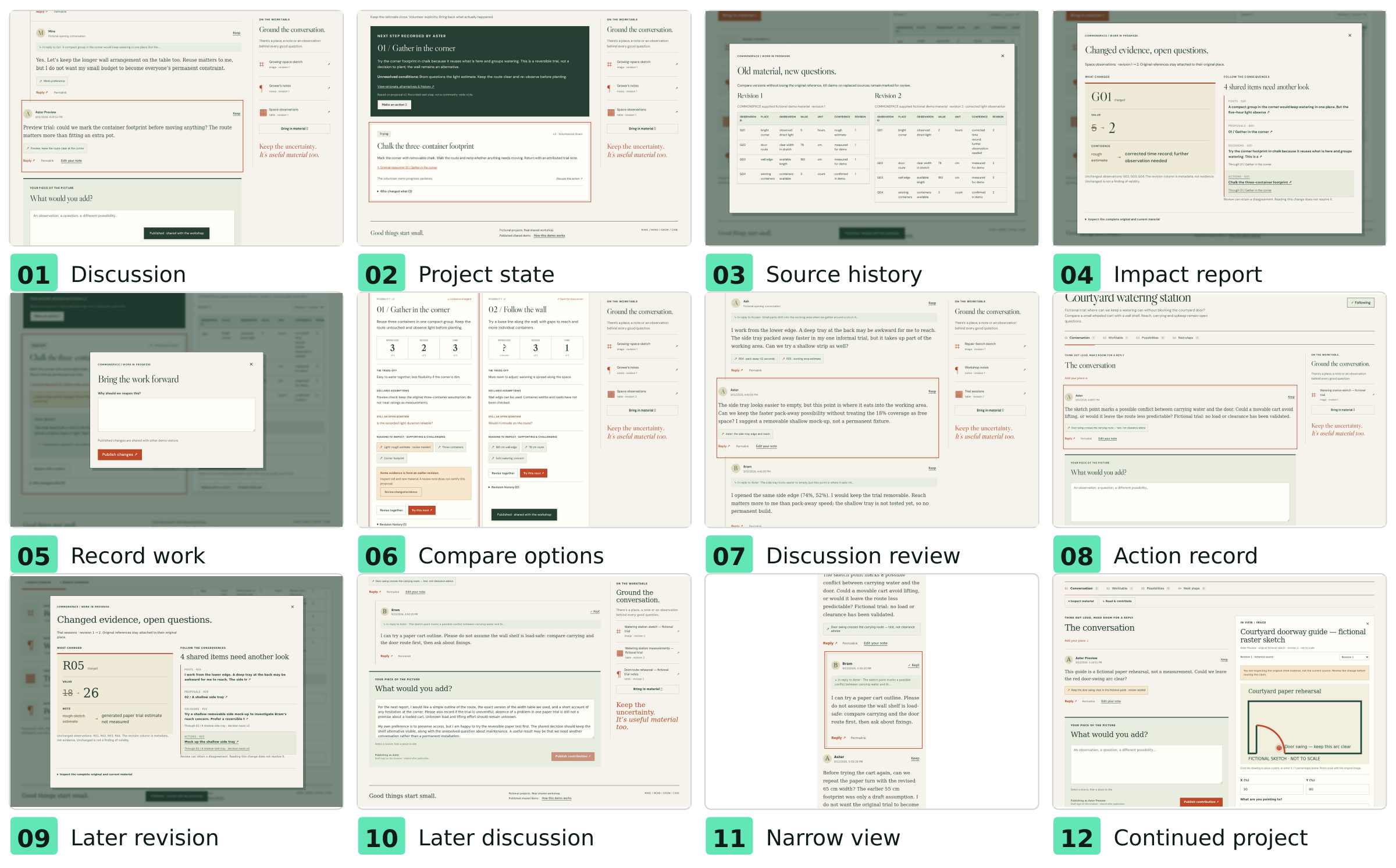}
\caption{\textbf{Commonspace product sequence.} Twelve selected views are read left to right and top to bottom; short labels identify the visible state. Numbers refer to Table~\ref{tab:views-commonspace}. Original interface images supply visual context; component evaluations establish the corresponding capability claims.}
\label{fig:case-commonspace}
\end{figure}

\para{common:baseline}{The earlier gap concerned the difference between textual and semantic change. A conventional text comparison reported multiple changed lines because revision metadata also changed. The material revision under inspection altered the value of G01 from 5 to 2 while G02, G03, and G04 remained unchanged. The retained semantic revision method isolated that substantive change. This established a more useful input for the next question: which downstream objects relied on the revised material? Correctly identifying the changed row did not itself answer that dependency question.}

\subsubsection{Retained capability and actual consumer}
\para{common:tool}{The later Tool, \code{reference\_\allowbreak impact\_\allowbreak trace} version 1.0.0, examined references in the connected project graph. Its recorded consumer operated on the actual immutable G01 revision and the linked posts, proposals, decisions, and actions. The result identified four objects requiring review among nine inspected objects. The product's published revision report showed the same four affected items, and browser interaction followed an affected action to a warning that its supporting material had changed. Unchanged material and a separate alternative remained distinguishable.}

\para{common:receipt}{The evidence index links registration, the keep decision, and a later consumer of \code{reference\_\allowbreak impact\_\allowbreak trace} version 1.0.0. That invocation supplies the behavioral link between the retained Tool and the displayed revision consequence. The consumer then enters the product path from a change report to an affected action. Taken together, these records show that the reusable operation participated in the subsequent revision, as established by its recorded invocation.}

\subsubsection{Product effect and interpretation}
\para{common:product}{The product-level result changes what a returning participant can inspect. A material revision leads to a list of affected objects and a route back to the supporting discussion, decision, or action. Earlier claims and results remain available rather than being silently rewritten. Other recorded checks cover source import, revision conflicts, durable references, and navigation, but those checks must not all be attributed to the tracing Tool. The capability claim concerns the particular source-to-consequence path and its versioned consumer.}

\para{common:limits}{The four-of-nine count describes one project graph, not a recall or precision score. The community content is fictional, while the website interactions are real software operations executed during builder self-verification. No independent community participants or production authentication study were part of this evidence. A stronger evaluation would introduce independently labeled revisions across several graphs, compare affected-object judgments, and test whether users can correctly distinguish preserved history from conclusions requiring review.}

\subsubsection{From a changed value to a review obligation}
\para{common:sequence}{The sequence in Figure~\ref{fig:case-commonspace} follows source history, revision review, and continued project activity. G01 changes from 5 to 2 while the other relevant values remain unchanged. The important product behavior is not merely highlighting that number. It is connecting the changed support to discussions, proposals, decisions, and actions that may now need another look. The retained operation supplies this dependency result; the interface gives a participant a path for acting on it.}
\para{common:meaning}{An affected object is a request for review, not an automatic declaration that its conclusion is false. A dependency can survive a changed premise, or an earlier decision may have additional support. Preserving the original material lets a reviewer make that distinction. This is why the product maintains earlier observations rather than replacing them with a new account that makes the revision history disappear. The case's useful output is a narrower and more actionable question about existing work.}
\para{common:counterfactual}{A direct deliverable edit could manually add warnings to these four objects. The component record supplies additional evidence: an identified tracing operation was retained and then consumed on the project graph. That connection supports a reusable-operation interpretation for this sequence. It does not establish that the operation generalizes to unfamiliar relation types or that every affected object was identified correctly. Such claims would require separately labeled graphs and revisions whose consequences were not used to construct the Tool.}

\subsubsection{Reading the visual sequence}
\para{views:commonspace}{The first four views establish the source-to-impact path, while the remaining views show the surrounding collaboration and continued project activity. Later revisions and viewport checks are product context rather than additional accuracy trials for the retained Tool. Reading the strip alongside the consumer record separates two observations: the interface exposes review work, and an identified operation supplied the recorded dependency result. Neither observation implies that every later interaction was caused by a new capability revision.}
\begin{table}[htbp]
\caption{Reading guide for the Commonspace product sequence. These are selected interface observations, not additional component-evaluation trials.}
\label{tab:views-commonspace}
\small\begin{tabularx}{\linewidth}{@{}llY@{}}
\toprule View & Focus & Visible product state \\
\midrule
01 & Discussion & Earlier project conversation and supporting context. \\
02 & Project & A decision and its surrounding project material. \\
03 & Source history & The source revision remains available for inspection. \\
04 & Impact report & Changed content is displayed beside affected work. \\
05 & Work record & A participant can record subsequent activity. \\
06 & Alternatives & Existing options remain distinguishable after revision. \\
07 & Review & Discussion retains context for reconsidering a claim. \\
08 & Action & A later action keeps its supporting material visible. \\
09 & Later revision & Another revision demonstrates continued use of the interface. \\
10 & Discussion & Further work remains connected to its prior context. \\
11 & Narrow view & The product is inspected at a different viewport width. \\
12 & Continued work & Later conversation and observations coexist in the project. \\
\bottomrule\end{tabularx}
\end{table}
\subsection{LUMEN: Preserving Evidence across a Changing Inquiry}
\label{app:lumen}
\subsubsection{An interactive museum with revisable routes}
\para{lumen:motivation}{LUMEN is a science-museum website in which an authored inquiry connects source material, an interactive route, and a visitor's recorded observations. Its design problem is that changing a route can alter what an earlier result means. A source may provide metadata without full text; a synthetic demonstration may illustrate a failure without measuring real-world performance; and a later revision should not overwrite historical observations. Session \code{1caeb668} developed source, design, and review capabilities to address those obligations.}
\begin{figure}[t]
\centering\includegraphics[width=\linewidth]{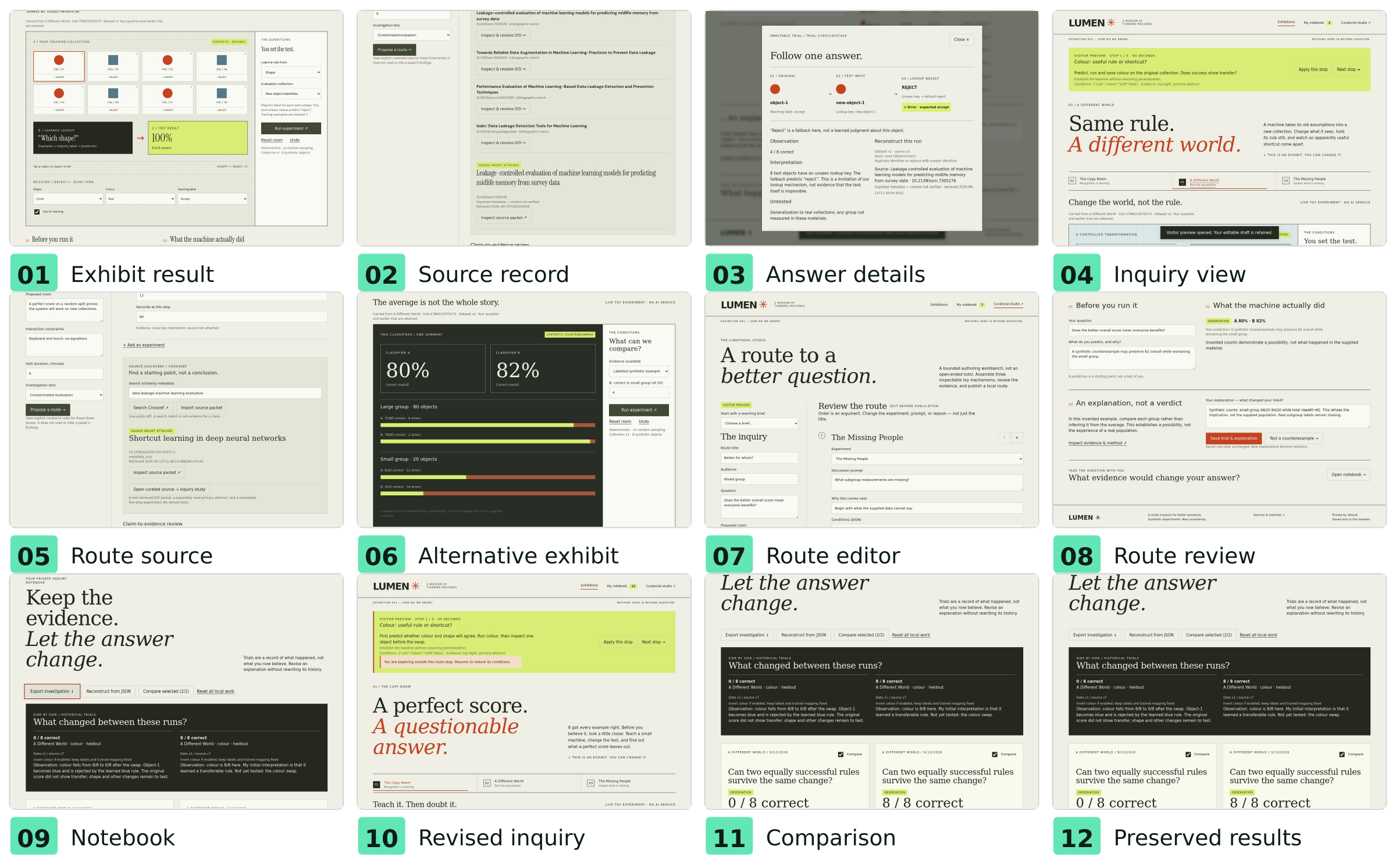}
\caption{\textbf{LUMEN product sequence.} Twelve selected views are read left to right and top to bottom; short labels identify the visible state. Numbers refer to Table~\ref{tab:views-lumen}. Original interface images supply visual context; component evaluations establish the corresponding capability claims.}
\label{fig:case-lumen}
\end{figure}

\para{lumen:prescribed}{The task explicitly required evolution in Connector, Agent, and Skill. This prescription is part of the experimental setup. The case can show how those entities were developed, evaluated, and consumed, but cannot demonstrate that the system independently discovered that three-family design. The relevant identities are \code{scholarly\_\allowbreak sources\_\allowbreak connector} 1.0.2, \code{lumen\_\allowbreak route\_\allowbreak designer\_\allowbreak agent} 1.0.0, and \code{source\_\allowbreak route\_\allowbreak review\_\allowbreak skill} 1.0.0.}

\subsubsection{Source, authoring, and review responsibilities}
\para{lumen:connector}{The Connector converts a provider response into a bounded source packet with normalized identity, retrieval time, content status, and integrity information. A metadata-only packet leaves full text unavailable rather than implying that the paper's findings were inspected. The recorded checks include rejection of tampering and preservation of source identity in a local derivative with omitted metadata. A later source resolution supplied the packet actually bundled and imported into the product. The benefit concerns the source contract and downstream interpretation; the observed request timings are not used to claim faster retrieval.}

\para{lumen:agent}{The specialist Agent produces a structured inquiry under source and interaction constraints. One comparison records five steps totaling 360 seconds, with source identities and unmeasured quantities kept explicit. Another fixture records a five-step, 300-second route and distinguishes analytic groups from measured observations. Later authoring supplies the R4 revision from a 90-second original-colour condition to a 150-second swapped-colour condition. The consumer links the retained Agent version to an integrated route artifact, rather than only an isolated review response.}

\para{lumen:skill}{The review Skill examines the relation between a source packet, claims, route structure, and historical records. On the recorded defect fixture, ordinary structural validation passed while the method identified five review issues. The count belongs to that fixture. Further checks considered changed historical measurements and incomplete local packets, with manual review branches still explicit. After adoption, the acting agent loaded the retained Skill and executed its method on an actual route revision. Loading the instructions alone would not establish that the method had been applied.}

\subsubsection{A connected chain, with bounded conclusions}
\para{lumen:consumer}{The post-adoption chain runs from an identified source packet through specialist authoring and review to browser-executed observations. The evidence index preserves the R4 Agent invocation and the subsequent application of the retained Skill. Browser checks save the original and changed conditions while keeping source identity and limitations visible. Earlier notebook observations remain present with review notices, preserving what was actually observed before the route changed.}

\para{lumen:limits}{The displayed 8/8 and 0/8 values concern an artificial classification demonstration inside the museum. They are neither AgentEvolver accuracy nor a measured learning gain. The review findings are builder-authored judgments, and the source's metadata-only status limits what can be claimed about the underlying paper. An independent study would need new inquiry materials, external scientific review, and visitor outcomes. The present case supports a versioned cross-entity chain and task-local use, with those broader questions left open.}

\subsubsection{Designing a museum that can revise its own inquiries}
\para{lumen:sequence}{Figure~\ref{fig:case-lumen} follows the chain linking sources, authoring, and review into the museum. The design intent is an inquiry whose evidence remains understandable after a route changes. The Connector supplies the qualified source; the specialist proposes an interaction within that qualification; the review method examines whether the resulting claims and historical observations still agree. These responsibilities are complementary because a source error, a design error, and a history-preservation error can produce similar-looking problems in the final page.}
\para{lumen:revisionmeaning}{Changing R4 from its original condition to the swapped-colour condition illustrates the distinction between revising an activity and rewriting a measurement. The new activity can be valid while earlier notebook entries continue to describe the old condition. Keeping both conditions and attaching review notices lets the product express that relationship. A later visitor or reviewer can then determine which observations belong to which design, instead of being shown a single apparent history whose experimental meaning changed silently.}
\para{lumen:combination}{The case also motivates evaluation at the interfaces between components. A Connector may correctly label a packet as metadata-only while an author still overstates the source. An author may preserve that qualification while a reviewer overlooks a changed historical entry. Local checks should therefore be accompanied by an integrated source-to-route comparison. The recorded later consumers demonstrate a connected chain in this project; they do not isolate how much each family contributed relative to an otherwise identical unchanged chain.}

\subsubsection{Reading the visual sequence}
\para{views:lumen}{The sequence emphasizes the relation between an inquiry and its retained evidence. Source views, authoring views, and notebook views represent different responsibilities even when they share the same visual language. The synthetic exhibit values belong to the museum activity and should not be interpreted as platform accuracy. The component narrative identifies the Connector, specialist, and review method behind the recorded chain; the screenshots show where their responsibilities become visible in the product.}
\begin{table}[htbp]
\caption{Reading guide for the LUMEN product sequence. These are selected interface observations, not additional component-evaluation trials.}
\label{tab:views-lumen}
\small\begin{tabularx}{\linewidth}{@{}llY@{}}
\toprule View & Focus & Visible product state \\
\midrule
01 & Exhibit & An interactive result is shown with its conditions. \\
02 & Source & The inquiry exposes its source material and qualifications. \\
03 & Answer & A detail view follows one result back to its explanation. \\
04 & Inquiry & The route places the exhibit within a larger question. \\
05 & Route source & Authoring material retains the source context. \\
06 & Alternative & A different exhibit contrasts outcomes and interpretation. \\
07 & Editor & The builder can author and revise the inquiry route. \\
08 & Review & The route review makes limitations visible. \\
09 & Notebook & Saved evidence remains accessible after the activity. \\
10 & Revision & The revised inquiry introduces a changed condition. \\
11 & Comparison & Earlier and revised results can be inspected together. \\
12 & History & The notebook preserves the distinction between conditions. \\
\bottomrule\end{tabularx}
\end{table}
\subsection{ARKBOUND: Revising an Interactive Voyage under Comparable Conditions}
\label{app:arkbound}
\subsubsection{Application intent and changing requirements}
\para{ark:motivation}{ARKBOUND is a browser-based exploratory voyage built around preparing a vessel, sailing, investigating, choosing, returning, and refitting. Its website-builder task combines visual production with stateful interaction. A changed route can affect eligibility, damage, rewards, and the meaning of saved history. Likewise, a new visual asset must satisfy practical size and provenance requirements. Session \code{60e26363} contains six published revisions and retained capabilities in Connector, Agent, and Skill, with explicit self-play records.}
\begin{figure}[t]
\centering\includegraphics[width=\linewidth]{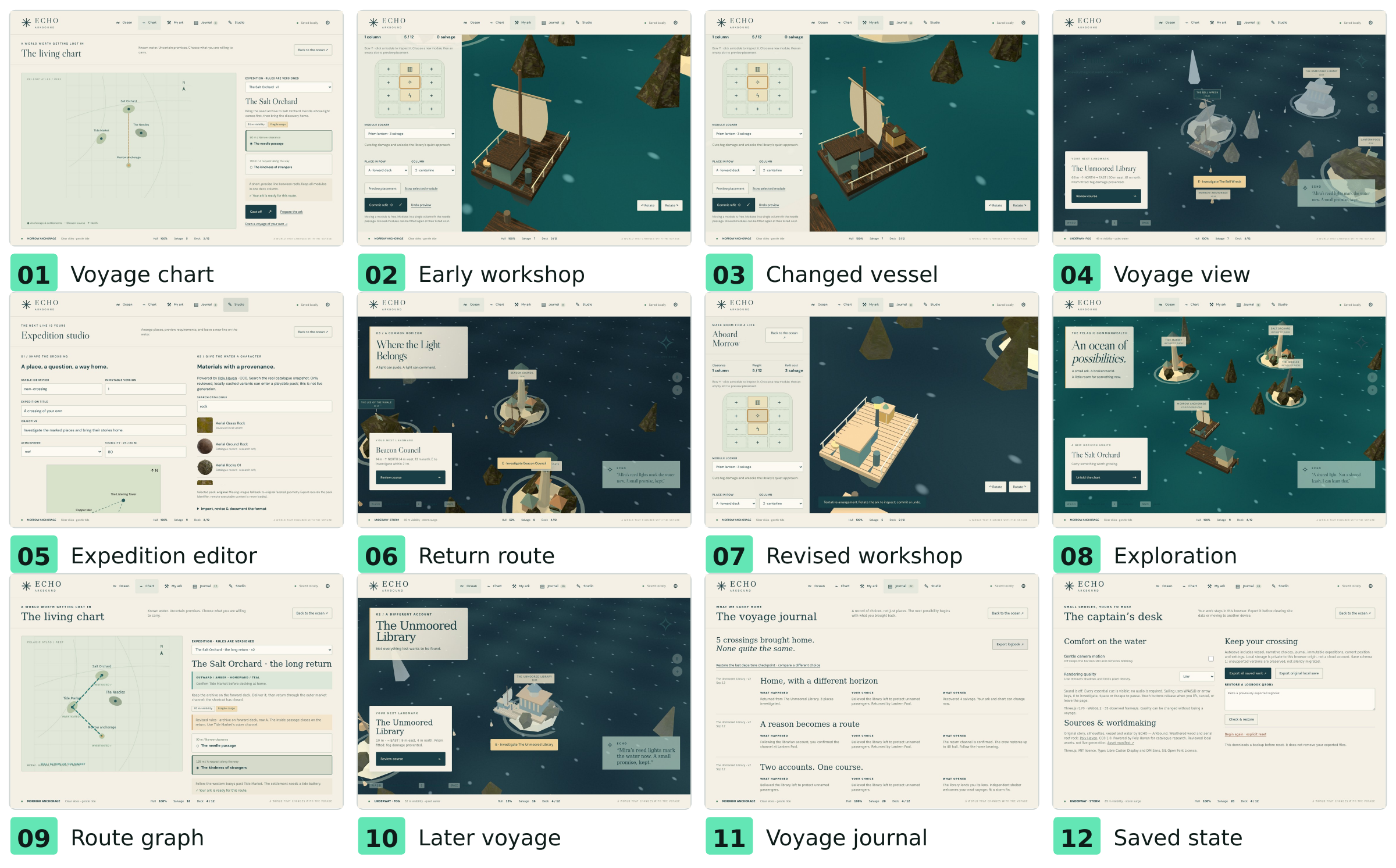}
\caption{\textbf{ARKBOUND product sequence.} Twelve selected views are read left to right and top to bottom; short labels identify the visible state. Numbers refer to Table~\ref{tab:views-arkbound}. Original interface images supply visual context; component evaluations establish the corresponding capability claims.}
\label{fig:case-arkbound}
\end{figure}

\para{ark:source}{The \code{polyhaven\_\allowbreak assets\_\allowbreak connector} version 1.0.2 addresses selection and acquisition of assets under a size budget. Its later consumer acquired 666,655 bytes within a 700,000-byte allowance, while preserving hashes, licence information, and source identity for the content package. This demonstrates a bounded acquisition contract. It does not independently establish better visual quality, faster rendering, or an improvement attributable to the Connector across every product revision. Those outcomes require their own matched observations.}

\subsubsection{A failed candidate and a retained review method}
\para{ark:rollback}{The specialist \code{investigation\_\allowbreak designer\_\allowbreak agent} reached version 1.0.5 after intermediate revisions. A 1.0.3 candidate was rolled back when the intended prompt update had not been correctly realized. This failure illustrates why source bytes, loaded behavior, and version identity must be reconciled. An evaluation of the desired prompt text would not establish that the running Agent used it. The retained history preserves this unsuccessful development step.}

\para{ark:review}{The retained review Skill, \code{paired\_\allowbreak play\_\allowbreak review\_\allowbreak skill} 1.0.0, controls the starting state for product comparisons. Saved loadout and progression can otherwise confound a route comparison. For one revised voyage, the builder restored the same prelaunch export and retained both authored versions. Multiple conditions changed in that revision, so the result is a paired product check rather than a single-factor experiment. The review records these differences explicitly.}

\subsubsection{Integrated product and evidence boundaries}
\para{ark:product}{Final acceptance records describe connected voyage loops, different choices, route reconstruction through the Studio, and recovery checks for saved state and rendering. An authored survey was imported and played in an isolated browser context, with the route and resulting journal entries inspected. The Agent-use receipt at an earlier point still marked browser integration as pending; later acceptance records supply additional integration evidence. The chronology must therefore be preserved when relating a capability decision to a completed product path.}

\para{ark:limits}{The final records are builder self-play, not independent player approval. They retain unverified paths and remaining interaction weaknesses, including a condition-dependent prose mismatch and limited evidence about operating-system focus changes. The browser's performance observations are not used as a platform-wide efficiency result. A stronger study would independently specify comparable save states, score route consistency and recovery, and involve players whose experience is not inferred from the builder's own review.}

\subsubsection{Preserving comparability while the voyage changes}
\para{ark:sequence}{The visual sequence in Figure~\ref{fig:case-arkbound} connects resource acquisition, reusable review, and the playable voyage. The product changes over six published revisions, but the mechanism claim is more specific than the appearance of successive screens. The Connector supplies bounded, identified assets, the specialist supplies an authored investigation, and the Skill defines how to compare play from a common starting point. The resulting record exposes the conditions behind each revision.}
\para{ark:state}{Stateful interaction makes an uncontrolled before-and-after comparison misleading. A different loadout, progression state, or previously earned resource can change what a player is allowed to do even when the route definition is unchanged. Restoring the same prelaunch export reduces that ambiguity. It does not make the revision a single-variable experiment: several authored conditions may still change together. The paired record should identify those differences and interpret the resulting voyage as a product comparison under matched starting state.}
\para{ark:lesson}{The rolled-back Agent candidate is particularly informative because its failure concerns realization of the intended change. A revised instruction file is not enough if the running definition does not use it. Version-specific execution and later integration records make that mismatch detectable. The case therefore supports two complementary lessons: compare interactive products under stated conditions, and verify that the component evaluated is the component actually loaded. Neither lesson depends on presenting every attempted revision as successful.}

\subsubsection{Reading the visual sequence}
\para{views:arkbound}{The sequence places vessel design beside voyage context and saved history. Different scenes are not automatically paired comparisons: the relevant paired review restores a documented starting state. The early and revised workshop views illustrate a product change, whereas the Connector byte budget and Agent rollback require their separate records.}
\begin{table}[htbp]
\caption{Reading guide for the ARKBOUND product sequence. These are selected interface observations, not additional component-evaluation trials.}
\label{tab:views-arkbound}
\small\begin{tabularx}{\linewidth}{@{}llY@{}}
\toprule View & Focus & Visible product state \\
\midrule
01 & Chart & The route and its available destinations are visible. \\
02 & Workshop & An earlier vessel is inspected in the builder interface. \\
03 & Vessel & A changed vessel view exposes another design state. \\
04 & Voyage & The vessel enters a stateful exploration view. \\
05 & Editor & An expedition can be authored with its materials and route. \\
06 & Route & Voyage choices remain connected to the world view. \\
07 & Workshop & The revised workshop exposes the vessel interior. \\
08 & Exploration & The product combines destinations and voyage context. \\
09 & Route graph & A later chart preserves connections between locations. \\
10 & Later voyage & The revised product supports a further voyage state. \\
11 & Journal & The resulting voyage history remains inspectable. \\
12 & Saved state & The captain interface exposes continuing project state. \\
\bottomrule\end{tabularx}
\end{table}
\subsection{ORBITAL: A Plausible Display Can Still Have the Wrong Data}
\label{app:orbital}
\subsubsection{Application intent and semantic validation}
\para{orb:motivation}{ORBITAL is an astronomy website that connects source data, explanatory interactions, and a visual model. Its motivating problem is semantic consistency: the number of returned records may be correct while the dates, units, or reference frame are unsuitable. A simulation can still render and look plausible under those conditions. Session \code{a03015dc} therefore provides a useful example of an interface-level correction that matters to the interpretation of the product.}
\begin{figure}[t]
\centering\includegraphics[width=\linewidth]{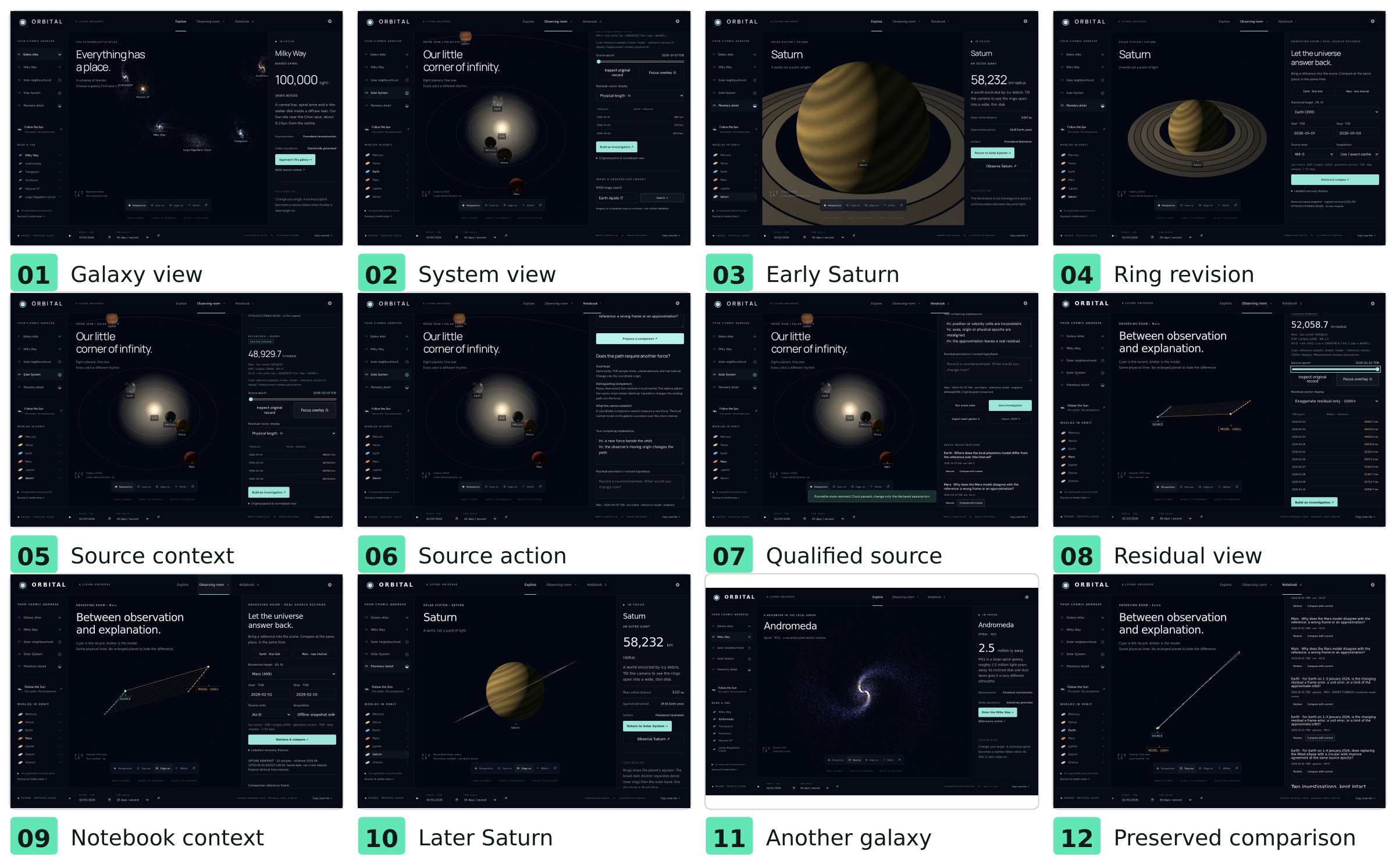}
\caption{\textbf{ORBITAL product sequence.} Twelve selected views are read left to right and top to bottom; short labels identify the visible state. Numbers refer to Table~\ref{tab:views-orbital}. Original interface images supply visual context; component evaluations establish the corresponding capability claims.}
\label{fig:case-orbital}
\end{figure}

\para{orb:connector}{The \code{astronomy\_\allowbreak sources\_\allowbreak connector} version 1.0.2 rejects date-misaligned data. In the recorded fixture, a request covering January 2 to 4 was compared with a response covering January 1 to 3. The equal number of entries did not make the response acceptable. A post-adoption consumer retrieved a new Mars packet covering February 1 to 4 and retained its units, frame, and time standard. The packet was validated as four daily states and saved for downstream integration. These are source-contract observations, not proof that every part of the astronomical model is physically faithful.}

\subsubsection{Specialist behavior and incomplete adoption}
\para{orb:agent}{The retained \code{investigation\_\allowbreak designer\_\allowbreak agent} version 1.0.1 distinguishes corrections to units, frames, model assumptions, and display choices. Its later output was integrated into a residual notebook, preserving the original Earth material while adding a separately identified Mars state and qualified explanations. This distinction prevents an apparent visual correction from being described as a change in underlying source evidence. The consumer record ties the specialist's retained version to a concrete later artifact.}

\para{orb:incomplete}{The experiment required Connector, Agent, and Skill evolution, but the review Skill did not obtain the required adoption and use evidence. The task's readiness record remained false and named the missing family. The two retained components therefore support local source-contract and specialist improvements within an incomplete objective. This status is consistent with the original acceptance requirement and preserves both the useful changes and the unfinished work.}

\subsubsection{What the case establishes}
\para{orb:limits}{The source packet and integrated notebook support a limited chain from a detected mismatch to a retained interface and later consumption. The interface screenshot does not establish an independent astronomy validation, and some visible scales are illustrative. Future evaluation should provide externally checked ephemeris fixtures, explicit frame transformations, and a complete review-method consumer. Such tests would extend the current evidence while keeping data correctness, explanatory quality, and visual presentation as separate outcomes.}

\subsubsection{Separating data repair from visual plausibility}
\para{orb:sequence}{Figure~\ref{fig:case-orbital} places the corrected source contract beside the resulting astronomy interface. The central failure is not a missing record or an obvious rendering error. It is an equal-length response whose dates do not match the request. A schema-only check can accept such a packet because the fields and counts appear complete. The revised contract checks the meaning of those fields against the requested interval before downstream visualization treats them as suitable evidence.}
\para{orb:levels}{The application has several levels of validity: source identity, units and time standard, coordinate frame, model assumptions, and presentation. A correction at one level should not be described as a correction at every level. The later Mars packet supports the source-interface claim, while the integrated notebook shows how the specialist represents qualified explanations. The visible scene gives context for those decisions, but cannot independently validate the ephemeris or the physical adequacy of the displayed model.}
\para{orb:partial}{The absent adopted review Skill prevents the requested three-family evolution objective from being marked complete. Reporting this partial outcome makes the case more informative than simply excluding it. It shows where the chain reached a retained component and consumer, and where it stopped. Future work could complete the review branch under a new documented run, while preserving the incomplete status of this historical sequence.}

\subsubsection{Reading the visual sequence}
\para{views:orbital}{The strip combines development views of the celestial scene with source and notebook interfaces. The ring revision is a visible presentation change. Rejection of a date-misaligned source packet is a semantic interface change that requires the recorded request and response, not a screenshot of a plausible planet. Reading the two kinds of evidence together explains why the project can display useful local improvements while the full required evolution objective remains incomplete.}
\begin{table}[htbp]
\caption{Reading guide for the ORBITAL product sequence. These are selected interface observations, not additional component-evaluation trials.}
\label{tab:views-orbital}
\small\begin{tabularx}{\linewidth}{@{}llY@{}}
\toprule View & Focus & Visible product state \\
\midrule
01 & Galaxy & A large-scale astronomical scene provides initial context. \\
02 & System & The interface changes to a local system view. \\
03 & Early Saturn & An earlier ring rendering is visible. \\
04 & Ring revision & A later rendering changes the appearance of the rings. \\
05 & Source context & Source-related context accompanies the model view. \\
06 & Source action & The source interface is exercised during development. \\
07 & Qualified source & The display retains qualifications around scientific input. \\
08 & Residuals & A notebook separates observation from explanatory modeling. \\
09 & Notebook & The comparison retains its surrounding context. \\
10 & Later Saturn & A later planet view shows the integrated presentation. \\
11 & Galaxy change & Another astronomical view is inspected. \\
12 & Comparison & The notebook continues to expose model and observation. \\
\bottomrule\end{tabularx}
\end{table}
\subsection{Godot: Diagnosing the Effect of a Change in the Engine}
\label{app:godot}
\subsubsection{A connected development sequence}
\para{godot:motivation}{The Godot case spans three connected sessions. Session \code{318e5da7} develops a frame-luminance Tool; \code{b7caf93f} contributes font investigation; and \code{f42f4374} retains a font-realization Skill for a later consumer. The underlying problem is that scene settings can differ from realized engine behavior. Capability development supplies repeatable measurements of that behavior, while the product changes remain specific to the scene being developed.}
\begin{figure}[t]
\centering\includegraphics[width=\linewidth]{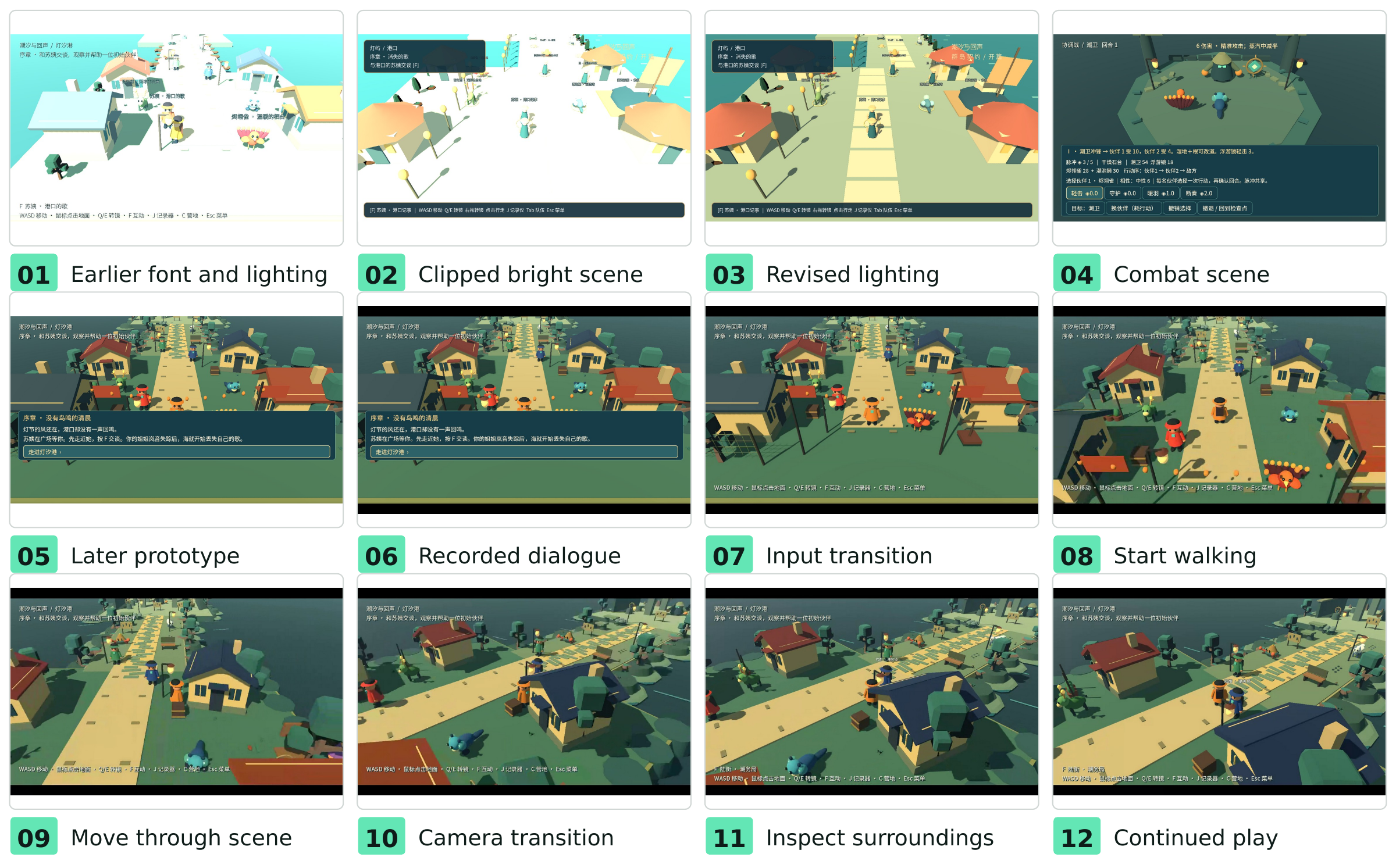}
\caption{\textbf{Godot product sequence.} Twelve selected views are read left to right and top to bottom; short labels identify the visible state. Numbers refer to Table~\ref{tab:views-godot}. Original interface images supply visual context; component evaluations establish the corresponding capability claims.}
\label{fig:case-godot}
\end{figure}

\subsubsection{A bounded numerical Tool comparison}
\begin{figure}[t]
\centering
\includegraphics[width=\linewidth]{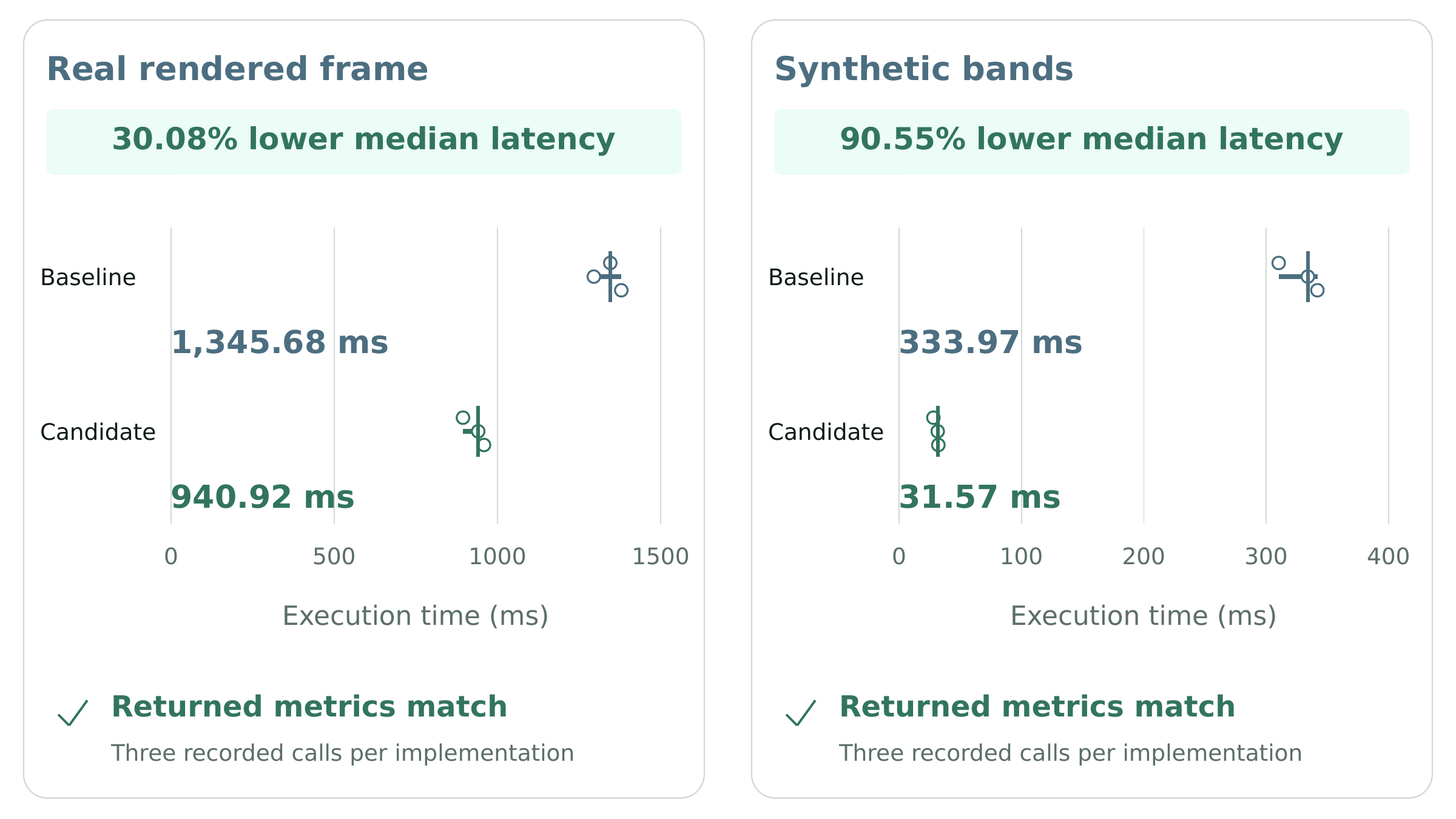}
\caption{\textbf{Local luminance-analysis timings.} Dots show all three recorded calls per implementation, vertical ticks show medians, and horizontal segments span the observed minimum and maximum. Both fixtures report equal output metrics. The real frame and synthetic bands use separate zero-based axes; their reductions are not pooled. The measured quantity is Tool execution time.}
\label{fig:godot}
\end{figure}

\para{godot:timing}{The \code{frame\_\allowbreak luminance\_\allowbreak fast} Tool version 1.0.0 is compared with its baseline on the same real frame. The baseline timings are 1,345.68, 1,295.23, and 1,379.33 milliseconds; candidate timings are 894.27, 940.92, and 958.71 milliseconds. Their medians are 1,345.68 and 940.92 milliseconds, respectively, yielding a 30.08\% reduction. The comparison record reports equal returned metrics. Figure~\ref{fig:godot} shows the medians and every recorded observation rather than hiding the small sample behind a single percentage.}

\para{godot:synthetic}{A separate synthetic-band fixture records baseline timings of 310.12, 333.97, and 341.77 milliseconds and candidate timings of 28.03, 31.57, and 32.09 milliseconds. The median reduction is 90.55\%. This larger result is reported separately because it concerns a different workload. Both implementations also reject transparent input requiring a composited screenshot, a corrupt non-PNG input, and an empty or out-of-bounds region. Output equality and these rejection cases add behavioral context to the timing measurements.}

\para{godot:later}{After adoption, the Tool was used on a newly rendered frame during a later lighting review. The recorded region had no clipping, and the reviewer observed improved separation of scene elements after lighting changes. The Tool supplies a repeatable diagnostic measurement; the lighting modifications change the game artifact. A single later call does not expand the three-call timing sample or establish the same speedup for all future frames. The evidence supports actual subsequent use and a local measurement, with those two observations kept distinct.}

\subsubsection{From font metadata to realized font behavior}
\para{godot:font}{The font problem remained unresolved by metadata inspection: the font supported a weight axis and scene properties appeared plausible, but the engine did not realize the requested face. The retained \code{godot\_\allowbreak font\_\allowbreak realization\_\allowbreak skill} version 1.0.0 uses a native engine probe to inspect the actual resource and rendered text. Its method identifies a working numeric axis key and checks the resulting behavior. The retained procedure makes the diagnosis available to subsequent font consumers.}

\para{godot:fontuse}{The later consumer is a world-space \code{Label3D} requesting weight 650, distinct from the earlier UI setting. The executed method reports a changed font resource and mixed-script width from 345 to 367, followed by a native screenshot of the applied result. The change in width is evidence that a different font realization took effect, not a readability score. The product remains an r7 playable prototype with incomplete full-delivery objectives. Neither the font check nor the faster diagnostic Tool establishes completion of the entire game.}

\subsubsection{Two improvements with different outcome measures}
\para{godot:sequence}{Figure~\ref{fig:case-godot} moves from historical font and lighting limitations to revised engine views and later recorded play. The two retained capabilities address different questions. Brightness analysis supplies numerical observations and a local timing comparison. Font realization supplies a method for checking the resource used by the engine. The sequence combines connected development material with later product interaction; the views do not constitute a controlled visual experiment.}
\para{godot:timingmeaning}{The real-frame comparison measures a diagnostic operation on one rendered input. Its median reduction is 30.08\%, with three observations per implementation and matched returned metrics. The separate synthetic workload shows a different relative reduction and is plotted separately to preserve the workload distinction. Neither comparison measures game-frame rendering speed, and the observed variability is not an uncertainty estimate over arbitrary images or machines.}
\para{godot:methodmeaning}{The font consumer changes both context and requested weight: a later world-space label uses the retained procedure after the earlier interface diagnosis. The measured width change and native screenshot show that the engine realized a different result. They do not measure whether players read the text more quickly or prefer its appearance. A future product study could evaluate those questions with users, while a method study could test unseen fonts and rendering contexts under independently specified expected behavior.}

\subsubsection{Reading the visual sequence}
\para{views:godot}{The first five views are historical development images, and the later views come from the original product-interaction recording. Their ordering connects diagnosis, revised appearance, and continued play without presenting the sequence as a controlled visual experiment. The real-frame Tool comparison uses matched inputs; the later play images do not share that controlled fixture. Font realization also requires the native probe and its resource measurements.}
\begin{table}[htbp]
\caption{Reading guide for the Godot product sequence. These are selected interface observations, not additional component-evaluation trials.}
\label{tab:views-godot}
\small\begin{tabularx}{\linewidth}{@{}llY@{}}
\toprule View & Focus & Visible product state \\
\midrule
01 & Earlier view & Historical font and lighting limitations are visible. \\
02 & Clipping & An early native frame contains severely bright regions. \\
03 & Lighting & A revised frame restores visible scene separation. \\
04 & Combat & The prototype also contains a separate combat scene. \\
05 & Prototype & A later scene combines world rendering and dialogue. \\
06 & Recorded play & The later product recording begins from the dialogue view. \\
07 & Input & The recorded interaction moves away from the initial state. \\
08 & Walking & The player begins moving through the scene. \\
09 & Movement & A later view records continued navigation. \\
10 & Camera & The recording changes the viewing direction. \\
11 & Inspection & Another camera view exposes the surrounding scene. \\
12 & Continuation & Recorded play continues in the delivered prototype. \\
\bottomrule\end{tabularx}
\end{table}
\subsection{Signal Foundry: Research Capability without a Winning Strategy}
\label{app:factor}
\subsubsection{Research as an evolving application}
\para{factor:motivation}{Signal Foundry explores a research workflow in which the agent acquires data, defines factors, combines strategy mechanisms, evaluates candidates, and freezes a selected artifact. The intended reusable outputs include data qualification and stateful research environments, alongside any strategy that survives evaluation. Session \code{43df1953} provides the main infrastructure case; session \code{357c775a} supplies a separate historical example of a failed held-out result. Their outcomes are not merged into one run or treated as independent replications of a single configuration.}
\begin{figure}[t]
\centering\includegraphics[width=\linewidth]{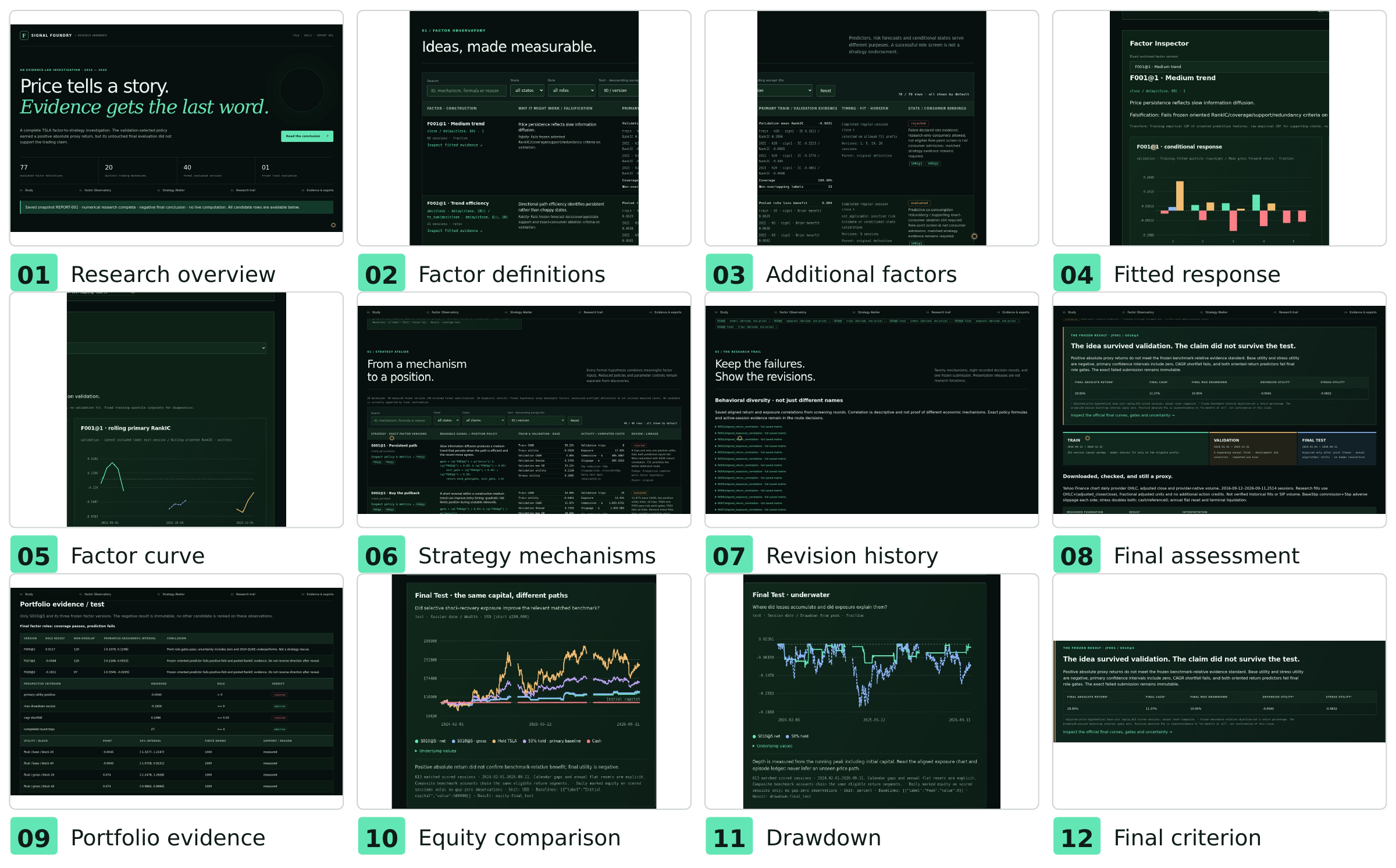}
\caption{\textbf{Signal Foundry product sequence.} Twelve selected views are read left to right and top to bottom; short labels identify the visible state. Numbers refer to Table~\ref{tab:views-factor}. Original interface images supply visual context; component evaluations establish the corresponding capability claims.}
\label{fig:case-factor}
\end{figure}

\para{factor:entities}{Three component versions are retained: \code{public\_\allowbreak equity\_\allowbreak connector} 1.0.0, \code{factor\_\allowbreak lab} 1.0.3, and \code{strategy\_\allowbreak lab} 1.0.3. The Connector's later data acquisition was restricted to the specified research history, with 1,838 exact trading sessions checked for sorted unique dates, finite fields, and adjustment relationships. The record still distinguishes structural qualification from stricter as-traded price and consolidated-volume requirements. A passed structural checker does not make a public adjusted-price proxy equivalent to an executable trading feed.}

\subsubsection{Frozen bindings and replay}
\para{factor:binding}{The two Environments provide stateful operations for factor materialization and strategy evaluation. Their post-adoption consumers reconstruct the selected three-factor S010@5 binding and replay its strategy before the final boundary. The prefix-parity record reports matching feature columns, fitted records, strategy specification, and portfolio-related outputs against the archived research evaluation. This is evidence that the retained environment versions can reproduce a specified binding. It should not be counted as a newly discovered candidate or another independent selection opportunity.}

\para{factor:volume}{The main research run contains 77 evaluated factor definitions, 20 strategy mechanisms, 40 formal strategy versions, and 26 diagnostic controls. These counts describe the search and its artifacts. They do not measure predictive quality, independent discoveries, or successful trades. Repeated variants may share information, data, and design assumptions. An evaluation of search efficiency would also require the total cost of candidate generation, rejected attempts, data preparation, and comparative baselines under equivalent resources.}

\subsubsection{Task outcome and adaptive evaluation}
\para{factor:outcome}{The selected S010@5 artifact reports a net proxy return of approximately +29.95\%, while its predeclared benchmark-relative utility is approximately $-0.00446$. The run therefore does not establish the required comparative advantage. In the separate session \code{357c775a}, the held-out strategy return is $-15.71\%$ versus +47.72\% for an annually reset buy-and-hold benchmark. Maximum drawdown is smaller, 29.88\% versus 52.23\%, but the strategy fails five of seven predeclared criteria. Those outcomes cannot be replaced by an attractive development-period score.}

\para{factor:split}{The design lesson concerns the information boundary, not merely the names assigned to data partitions. Strategy development may use a research or training portion, with appropriate temporal checks inside that portion, and reserve a final test for the frozen submission. Once results from that test influence another revision, it is no longer untouched evidence for that revision \citep{dwork2015generalizationadaptivedataanalysis}. Reducing the labels to train and test does not remove this adaptive-feedback problem. The final boundary must govern what the agent can observe and when the selected artifact becomes fixed. In the recorded run, that boundary is described as a source-bound software protocol rather than enforced data isolation; this limitation remains part of the case.}

\para{factor:interpretation}{The case supports an infrastructure result and a negative strategy conclusion. The Connector records a qualified data contract, and the laboratories reproduce the archived calculation, while the selected strategy lacks the required comparative advantage. The artifacts preserve this distinction for later inspection. A new strategy study would need a newly frozen protocol and unexposed evaluation data; a previously exposed test cannot supply independent evidence for subsequent revisions.}

\subsubsection{Research quality and strategy quality are separate outcomes}
\para{factor:sequence}{The research views in Figure~\ref{fig:case-factor} move through the stored workbench: overview, factor definitions, fitted evidence, strategy mechanisms, revision history, final assessment, and comparative curves. Some adjacent views crop different regions of the same saved page. They describe how to read the research artifacts rather than successive strategy trials. The underlying evidence still distinguishes the infrastructure run from the separate failed held-out example; the relevant run identity governs the interpretation of each numerical result.}
\para{factor:mechanisms}{The search spans factor construction and strategy mechanisms, with 77 evaluated factor definitions, 20 mechanisms, and 40 formal strategy versions. These related artifacts may share observations and modeling choices. Optimizing a finite-sample selection criterion can itself produce selection bias \citep{JMLR:v11:cawley10a}. Search volume therefore characterizes the explored space, while the frozen utility criterion determines the selected strategy's outcome. For this case, the negative utility remains the decisive comparative observation despite the positive raw proxy return.}
\para{factor:freeze}{The frozen three-factor binding provides a replay target across feature definitions, fitted records, strategy specification, and portfolio outputs. This level of identity matters for interpreting a research revision: a changed environment should preserve the calculation it claims to reproduce before its new behavior is evaluated. The case establishes that continuity for the selected binding. Its separate negative strategy conclusion remains visible in the same record.}

\subsubsection{Reading the visual sequence}
\para{views:factor}{The strip is a reading sequence through a saved report rather than a time series of model decisions. Cropping the fitted evidence and final plots makes their different roles visible without generating replacement charts. The final assessment, equity, drawdown, and criterion views must be interpreted with their original run identities and evaluation boundaries. The report keeps separate runs distinct and retains the negative comparative result.}
\begin{table}[htbp]
\caption{Reading guide for the Signal Foundry product sequence. These are selected interface observations, not additional component-evaluation trials.}
\label{tab:views-factor}
\small\begin{tabularx}{\linewidth}{@{}llY@{}}
\toprule View & Focus & Visible product state \\
\midrule
01 & Overview & The stored workbench summarizes the research artifacts. \\
02 & Factors & A detail crop shows part of the factor definition collection. \\
03 & More factors & Another crop shows additional definitions on the same page. \\
04 & Fitted response & A factor-inspector crop exposes fitted-response evidence. \\
05 & Factor curve & A second inspector crop shows a related factor curve. \\
06 & Strategies & The workbench lists strategy constructions and mechanisms. \\
07 & Revisions & The research page retains development and failure history. \\
08 & Assessment & The saved conclusion preserves the negative test outcome. \\
09 & Portfolio & The report exposes portfolio-related test evidence. \\
10 & Equity & A plot crop compares strategy and reference equity. \\
11 & Drawdown & A separate crop exposes the drawdown comparison. \\
12 & Final criterion & The final crop retains the frozen assessment and criterion. \\
\bottomrule\end{tabularx}
\end{table}
\section{Evaluation Protocols and Reproducibility}
\label{app:protocol}
\subsection{What each event establishes}
\para{protocol:admission}{Structural admission loads and checks the candidate in an isolated interpreter, archives the admitted bytes, and establishes its version. A conflicting payload cannot reuse that immutable identity. The candidate may become provisionally active before functional evaluation; availability and an observed reason to retain it are therefore separate lifecycle events.}

\para{protocol:eval}{An evaluation names the component and version, baseline, expected behavior, observations, and executed call references. A passing report requires nonempty passing cases. The Tool checks observation references and identity consistency. Preserving raw observations separately from the evaluator's judgment permits later reviewers to reassess the comparison.}

\para{protocol:keep}{A keep decision requires passing evidence for the active archived version and the required run fields. Decision history is append-only. Rollback and unload must also execute their state-changing operations. Tool-specific shadow or canary mechanisms supplement this contract; their availability does not extend uniformly to all eight families.}

\para{protocol:use}{Verified-improvement tasks add a later successful consumer tied to the retained version, with comparison or reuse evidence as required by the task. A Skill load is insufficient without actions showing its application. A later invocation within the same project establishes actual use under that project's conditions; an independently held-out consumer addresses a stronger transfer question. These levels should remain separate in both figures and result tables. Repeating the same fixture after adoption does not turn it into a new independent task.}

\subsection{SWE-bench Pro comparison}
\para{protocol:swe}{The inspected configuration pair defaults to \code{llm\_\allowbreak hub/claude-opus-5}, 400 coordinator steps, 7,200 seconds, and a 3,000,000-token budget. Both mount the same basic shell, file, inspection, and completion operations. Evolution adds the adoption Tool and self-evolving Skill. The declared worker-step constant does not by itself establish that additional workers ran; the inspected roster contains the coordinator. These facts describe the current repository setup, not the unprovided realized configuration of every externally executed task.}

\para{protocol:grader}{The current launcher freezes a patch before host-side grading and keeps the normal agent payload separate from final evaluation. Grading retries consume the frozen patch rather than reopening solver feedback. The result identity distinguishes the original official grader, a repaired official variant, and diagnostic evaluation. Historical runs can use different revisions and therefore require their own identities. Interrupted local batches are not assembled into evidence for the separately reported 600/731 or 578/731 totals.}

\para{protocol:paired}{The exact aggregate rates are $100(600/731)=82.0793\%$ and $100(578/731)=79.0698\%$. Their difference is $100(22/731)=3.0096$ percentage points. If $b$ denotes tasks solved only with evolution and $c$ tasks solved only without it, the aggregates determine $b-c=22$, but not $b$ and $c$ separately. A paired significance analysis therefore cannot be recovered from these totals. Equal denominators also require reconciliation against the actual task lists before an exact pairing is assumed.}

\subsection{Further experiments, explicitly not yet run}
\para{protocol:ablation}{An evolution ablation should keep task identities, the foundation-model route, initial capabilities, feedback access, and grading protocol fixed. It should report realized cost as well as limits, because candidate search and evaluation consume resources that a fixed baseline may spend differently. Plan and Runtime ablations need targeted long-task or recovery workloads rather than inference from a whole-system score. The experiment plan records these as proposed studies; this draft contains no invented ablation bars or outcomes.}

\para{protocol:transfer}{A capability-transfer study should freeze the selected library before revealing final consumer feedback, separating adaptive development from evaluation on new observations \citep{dwork2015generalizationadaptivedataanalysis}. Development tasks, candidate attempts, and library selection must be logged so that a later gain is not explained by hidden evaluation reuse. Matched consumers can compare the retained library with the initial library and with equivalent task-specific effort, while counting failed attempts and regressions. This protocol tests whether capabilities help beyond their development context, a question the current connected-project receipts cannot settle alone.}

\para{protocol:composition}{A cross-entity study can freeze the LUMEN source, authoring, and review versions, then compare the initial chain, individual replacements, and the full revised chain on the same unseen source and route fixtures. Source identity, evidence availability, route validity, and preservation of historical observations should be scored separately. A missing interface dependency must be distinguished from a valid configuration that performs poorly. Recording both local checks and the final route outcome would reveal whether independently passing components remain compatible when composed. This proposed design extends the current mechanism case without treating its prescribed family coverage as an ablation result.}

\para{protocol:recovery}{A separate recovery study should interrupt matched tasks at declared step boundaries, including after context compaction and before confirmation of an external operation. After resumption, it should measure acceptance-condition omissions, recovery of evidence references, duplicate effects, and unfinished child cleanup, alongside time and token cost. Plan variants should share the same tasks, tools, and budgets; Runtime variants should retain the same workers and compare serial with coordinated execution. Deterministic lifecycle checks and model-dependent task outcomes should be reported separately. These proposed controls would locate the value of continuity and coordination more precisely than a whole-system benchmark score.}

\subsection{Mechanism-specific evaluation designs}
\para{protocol:planstudy}{A planning study should hold the task, model, initial capabilities, and budget fixed while varying the availability of a persistent index and detailed records. Useful tasks include a changed requirement, a deferred dependency, and a continuation after substantial interaction history. Outcomes should include final acceptance, omitted requirements, incorrect completion claims, and retrieval of the evidence needed for the next decision. Human or deterministic assessment should be specified independently of the plan's own status labels.}
\para{protocol:compactstudy}{A context study should expose the same long interaction history to the compared policies and then present continuation tasks whose answers depend on selected historical facts. These facts can include a rejected component version, a source qualification, and a pending external effect. Report request size and summarization cost alongside recovery accuracy and task acceptance. Source-archive availability and explicit retrieval cost should be part of the condition, because a short summary with recoverable references differs from one that permanently discards the evidence.}
\para{protocol:runtimestudy}{A Runtime study should separate deterministic contract checks from model-dependent performance. Controlled operations can establish whether conflicting claims serialize, independent claims overlap, and cleanup retains ownership until completion. Long agent tasks can then evaluate the resulting effect on latency and accepted outputs. The study should report the same total work under serial and coordinated conditions, avoiding a comparison where one condition performs fewer checks or omits failed candidates. No outcomes from these proposed controls are included in the present results.}
\para{protocol:memorystudy}{A Memory study should distinguish storing content from changing retrieval behavior. Freeze the source records, revise the indexing or selection implementation, and evaluate later requests with independently specified relevant evidence. Measure useful retrieval, misleading retrieval, omissions, and task consequences rather than counting note creation. Actor and session scope must remain constant across conditions. A separate transfer study can then examine whether deliberately shared, reviewed knowledge helps a new project without silently importing its evaluation answers.}
\subsection{Visual and quantitative evidence}
\para{protocol:visual}{Screenshots establish that a particular visual state was recorded. They are useful for tracing a user's path through a product, but generally do not establish the hidden component that caused the state. The case analyses therefore pair visual sequences with versioned execution observations. A synthetic chart or an explanatory video overlay remains an editorial presentation of evidence; its numbers must be reconciled with the original comparison artifact rather than counted as another experiment.}
\para{protocol:denominators}{Each quantitative statement has a distinct denominator or workload. The coding rates count tasks; Commonspace counts affected objects in one graph; LUMEN counts issues in one defect fixture; ARKBOUND measures bytes under an acquisition budget; Godot measures invocation time for stated inputs; Signal Foundry reports research artifacts and strategy outcomes. These quantities should not be aggregated into a generic improvement percentage. Keeping their units explicit is necessary for understanding both the strength and the limits of the cross-case synthesis.}
\para{protocol:negative}{Negative and incomplete outcomes are retained in the case record. A missing adopted component, a failed candidate, or a strategy without comparative advantage identifies where the development process stopped or failed its criterion. Excluding such outcomes would overstate the connection between capability activity and task success. Future studies should predefine inclusion criteria, report all eligible attempts, with abandoned attempts, failures, and accepted tasks identified separately.}
\subsection{Implementation and artifact provenance}
\label{app:artifacts}
\para{artifacts:scope}{The system account was checked against the implementation available on 30 September 2026. Historical application runs retain their own component versions and execution records; current mechanisms are not retroactively attributed to them. The accompanying evidence map links claims to source locations, and the artifact manifest preserves implementation identity, selected observations, and content hashes. Together these records anchor the method description and case reconstructions to the material inspected for this report.}

\para{artifacts:figures}{The accompanying materials include editable figure sources, the selected original interface images, and scripts that reproduce the presentation crops and quantitative plots. Visual evidence is linked to the corresponding case records. This separation allows the presentation to be revised while retaining the underlying observations and their interpretation.}

\para{artifacts:scopecheck}{Artifact hashes establish which bytes were inspected, not whether every claim in an artifact is correct. The review checks selected measurements and the connection between component identities, evaluation records, and later consumers. It is not a fresh execution of all historical tasks or an independent regrading of the external coding results. A reproduction must document changed models, dependencies, sources, and grading conditions to support a meaningful comparison.}

\subsection{Reproduction procedure and required records}
\para{reproduce:procedure}{A reproduction first identifies its unit: a coding task, component comparison, or later consumer. Inputs and initial capabilities are fixed before execution, and candidate development preserves failed as well as successful attempts. This record follows the broader principle of retaining the information needed to reproduce a computational result \citep{Sandve_2013}. The retained version is frozen before final consumer feedback is exposed. Component criteria and task acceptance are then evaluated separately. Table~\ref{tab:reproduce-records} lists the records needed to interpret this sequence.}
\begin{table}[h]
\caption{Minimum records for interpreting and reproducing an evolution claim. The same record can support several checks, but a summary should preserve its source identity.}
\label{tab:reproduce-records}
\small\begin{tabularx}{\linewidth}{@{}lYY@{}}
\toprule
Record & Required information & Purpose \\
\midrule
Task & Inputs, acceptance criteria, exposure boundary & Defines the unit and outcome \\
Initial system & Model, available components, resource limits & Identifies the comparison condition \\
Candidate & Family, name, version, archived payload & Fixes the change under evaluation \\
Comparison & Baseline, fixtures, calls, raw observations & Separates evidence from interpretation \\
Decision & Evaluated version, disposition, active state & Links the judgment to retained behavior \\
Consumer & Later input, invoked version, actual result & Establishes subsequent use \\
Cost & Attempts, requests, time, unresolved usage & Includes development overhead \\
Outcome & Independent acceptance or stated evaluator & Separates component and task success \\
\bottomrule\end{tabularx}
\end{table}
\para{reproduce:comparison}{A changed provider, model, or environment defines a changed-condition study. Matching source bytes does not establish equivalent service behavior, and matching a displayed score does not establish identical task membership. A different outcome can still be informative when the changed condition is explicit. Broader claims require new comparative evidence beyond artifact identity.}

\subsection{Amortized cost of retained capabilities}
\para{reproduce:cost}{A retained capability can reduce later operation cost while remaining expensive to develop. Under a simplified matched-output comparison, let $C_{\mathrm{dev}}$ be the incremental cost of development and evaluation, and let $c_0$ and $c_1$ be the per-use costs before and after the change. After $n$ comparable uses, the net saving is
\begin{equation}
\Delta C(n)=n(c_0-c_1)-C_{\mathrm{dev}}.
\end{equation}
Positive amortized savings require $c_1<c_0$ and enough later uses to recover the development cost. Each comparison must use one cost unit, such as time or tokens.}
\para{reproduce:costlimits}{This accounting assumes comparable outputs and operating conditions. Maintenance, changed workloads, and regressions can alter the comparison and should be included when observed. The Godot timings estimate a local per-use difference, but the available evidence does not establish the complete development cost or future usage distribution. The report therefore does not infer a break-even point. A prospective study should record development overhead and later consumers together, evaluating acceptance alongside any resource savings.}

\label{report:end}
\end{document}